\documentclass[10pt,twocolumn,letterpaper]{article}

\usepackage{iccv}              % To produce the CAMERA-READY version
\definecolor{iccvblue}{rgb}{0.21,0.49,0.74}
\usepackage[pagebackref,breaklinks,colorlinks,allcolors=iccvblue]{hyperref}
\usepackage{comment}
\usepackage{amssymb}
\usepackage{graphicx}
\usepackage{multirow}
\usepackage{threeparttable}
\usepackage{pifont} 
\usepackage{booktabs}

\def\paperID{8324} % *** Enter the Paper ID here
\def\confName{ICCV}
\def\confYear{2025}

\title{NavM\includegraphics[height=1.1ex]{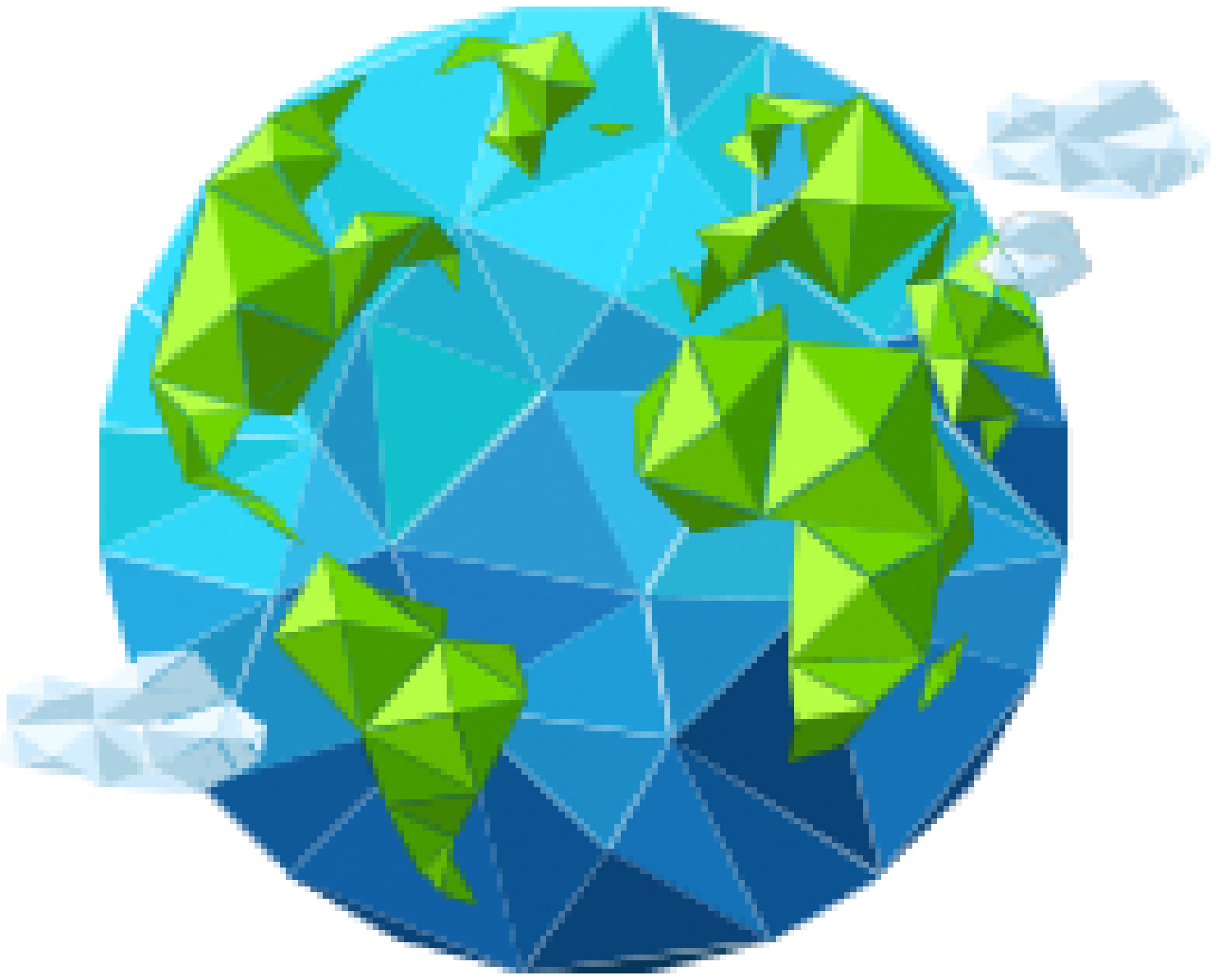}rph: A Self-Evolving World Model for Vision-and-Language Navigation in Continuous Environments}
\author{
    Xuan Yao\(^{1,2}\), Junyu Gao\(^{1,2}\), and Changsheng Xu\(^{1,2,3}\) \\
    \(^{1}\)State Key Laboratory of Multimodal Artificial Intelligence Systems (MAIS), \\
    Institute of Automation, Chinese Academy of Sciences (CASIA) \\
    \(^{2}\)School of Artificial Intelligence, University of Chinese Academy of Sciences (UCAS) \\
    \(^{3}\)Peng Cheng Laboratory, ShenZhen, China \\
    \texttt{yaoxuan2022@ia.ac.cn; \{junyu.gao, csxu\}@nlpr.ia.ac.cn}
}
\begin{document}
\maketitle
%\begin{abstract}
%Vision-and-Language Navigation in Continuous Environments (VLN-CE) requires agents to execute sequential navigation actions in complex environments guided by natural language instructions. Current approaches often struggle with generalizing to novel environments and adapting to ongoing changes during navigation.
%Inspired by human cognition, we present NavMorph, a self-evolving world model framework that enhances environmental understanding and decision-making in VLN-CE tasks. NavMorph employs compact latent representations to model environmental dynamics, equipping agents with foresight for adaptive planning and policy refinement. By integrating a novel Contextual Evolution Memory, NavMorph leverages scene-contextual information to support effective navigation while maintaining online adaptability. Extensive experiments demonstrate that our method achieves notable performance improvements on popular VLN-CE benchmarks. Code is available at \href{https://github.com/Feliciaxyao/NavMorph}{this https URL}.
%\end{abstract}

\begin{abstract}
	Vision-and-Language Navigation in Continuous Environments (VLN-CE) requires agents to execute sequential navigation actions in complex environments guided by natural language instructions. Current approaches often struggle with generalizing to novel environments and adapting to ongoing changes during navigation.
	Inspired by human cognition, we present NavMorph, a self-evolving world model framework that enhances environmental understanding and decision-making in VLN-CE. NavMorph employs compact latent representations to model environmental dynamics, equipping agents with foresight for adaptive planning and policy refinement. By integrating a novel Contextual Evolution Memory, NavMorph leverages scene-contextual information to support effective navigation while maintaining online adaptability. Extensive experiments demonstrate that NavMorph achieves notable performance improvements on popular VLN-CE benchmarks. Our Code is available at \url{https://github.com/Feliciaxyao/NavMorph}. %\href{https://github.com/Feliciaxyao/NavMorph}{this https URL}.
\end{abstract}
    
\vspace{-4mm}
\section{Introduction}\label{sec:intro}
\vspace{-2mm}
In recent years, Embodied AI~\cite{wang2024embodiedscan,mu2024embodiedgpt,majumdar2024openeqa,yang2024holodeck} has emerged as a pivotal research direction, attracting significant attention across computer vision, natural language processing, and robotics communities due to its interdisciplinary nature.
Among its core challenges, Vision-and-Language Navigation (VLN)~\cite{anderson2018vision} has gained prominence for its broad applicability in many intriguing real-world applications, such as robotic assistance~\cite{balatti2024robot,kuribayashi2024memory}, autonomous navigation~\cite{panda2023agronav,kayukawa2023enhancing}, and smart home systems~\cite{vanus2023using,toupas2023framework}.

%stands as a fundamental challenge, requiring agents to interpret natural language instructions, process visual information in dynamic 3D environments, and execute sequential actions to navigate toward target locations.comprehend visual inputs and natural language instructions, enabling them to model their surroundings and infer spatial structures of the environment. This capability is particularly crucial for path planning, especially in complex navigation scenarios involving multiple or diverse room configurations.

VLN tasks require agents to interpret natural language instructions, process visual information in dynamic environments, and execute sequential actions to reach target locations. To navigate effectively, agents must dynamically build a structured understanding of scenes from current observations and instructions, enabling them to predict the outcomes of their actions and make reasonable inferences about subsequent steps in response to ongoing environmental changes. 
Inspired by remarkable human capacity to construct mental representations of environments, facilitating deliberate planning for complex tasks like motor control, imagery, inference, and decision-making~\cite{Hao2023ReasoningWL}, world models have emerged as promising frameworks that explicitly model environmental dynamics by simulating actions and their effects on world states.
By maintaining a compact latent representation, world models encapsulate dynamic and structural characteristics through agent-environment interactions~\cite{lecun2022path}, enabling prediction of state transitions and potential future observations.
%This capability for internal simulation, referred to as `learning in imagination'~\cite{hafner2019dream}, can empower VLN agents to dynamically refine their environmental understanding and navigation strategies, substantially reducing the dependence on trial-and-error exploration during navigation episodes.
\begin{figure}
	\centering
	\includegraphics[width=0.98\linewidth]{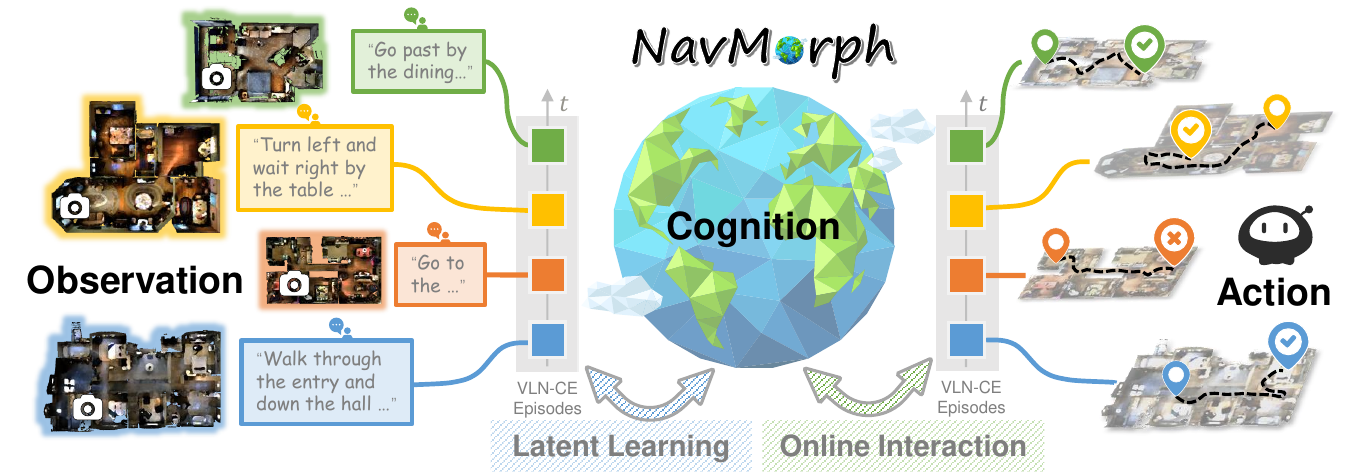}
	\vspace{-2mm}
	\caption{Self-Evolving World Model for Online VLN-CE tasks.}
	\label{fig:motivation}
	\vspace{-6mm}
\end{figure}
Despite a few pioneering efforts, such as Navigation World Models~\cite{Bar2024NavigationWM}, DreamWalker~\cite{Wang2023DREAMWALKERMP} and PathDreamer~\cite{koh2021pathdreamer}, have demonstrated the potential of world models for navigation, their application in VLN remains underexplored.
%Despite a few pioneering efforts demonstrating the potential of world models in navigation tasks, such as Navigation World Models~\cite{Bar2024NavigationWM}, DreamWalker~\cite{Wang2023DREAMWALKERMP} and PathDreamer~\cite{koh2021pathdreamer}, their application in VLN remains underexplored.
Specifically, these approaches either lack the capacity to learn underlying action-state transitions or rely on discrete state dynamics, limiting their ability to capture the continuous nature of spatial-temporal dynamics inherent in VLN tasks.
The limitation becomes even evident during online deployment, where agents frequently encounter complex and varied environments with substantial distributional shifts from training data~\cite{park2023visual,wu2024vision}. 
Such environmental disparities pose risks of performance degradation~\cite{gao2024fast}, while existing world model approaches~\cite{koh2021pathdreamer, Wang2023DREAMWALKERMP, Bar2024NavigationWM} focus primarily on the pre-training stage and lack mechanisms for continuous updating, therefore limiting their ability for online adaptation to novel environments.
%during online testing, thereby struggling to characterize novel environments or accumulate previous navigational experience effectively.
%necessitating immediate adaptation of VLN models while capturing ongoing dynamics and visual-language cues for precise decision-making.
%Such environmental disparities pose risks of performance degradation~\cite{gao2024fast}, necessitating immediate adaptation of VLN models while capturing ongoing dynamics and visual-language cues for precise decision-making.
%Existing world models in VLN~\cite{koh2021pathdreamer, Wang2023DREAMWALKERMP} primarily focus on pre-training stages and lack mechanisms for dynamic updating during online testing, thereby struggling to characterize novel environments or accumulate previous navigational experience effectively.
%Despite the promising potential, the application of world models for in VLN remains underexplored. Pioneering efforts such as PathDreamer~\cite{koh2021pathdreamer} and DreamWalker~\cite{Wang2023DREAMWALKERMP} have demonstrated initial success in incorporating world models into navigation.
%Despite a few pioneering efforts demonstrating the potential of world models in navigation tasks, such as Navigation World Models~\cite{Bar2024NavigationWM}, DreamWalker~\cite{Wang2023DREAMWALKERMP} and PathDreamer~\cite{koh2021pathdreamer}, their application in VLN remains underexplored.
%PathDreamer employs generative adversarial networks to synthesize visual predictions of unobserved views, while DreamWalker utilizes Monte Carlo Tree Search (MCTS) for discrete action planning.

To address the above issue, we propose NavMorph, a self-evolving world model framework tailored for VLN-CE tasks. As shown in Figure~\ref{fig:motivation}, NavMorph constructs a continuously evolving latent space that transforms vision-language inputs (observations) into effective navigation decisions (actions) through online interaction. 
%bridges observations and actions through cognitive modeling, where a continuously evolving latent space transforms visual-language inputs into effective navigation decisions through online interaction.
We design a tailored Recurrent State-Space Model (RSSM)~\cite{hafner2019learning} for VLN, which can capture continuous navigational dynamics by explicitly modeling latent action-state transitions.
%employs the Recurrent State-Space Model (RSSM)~\cite{hafner2019learning} to capture continuous navigational dynamics within VLN episodes by modeling latent action-state transitions. 
Extending this foundation, NavMorph incorporates self-evolution by progressively enhancing its latent representations, enabling agents to accumulate contextual information.
%for comprehensive environmental understanding and effective adaptation to novel scenes.
%Our world model employs the Recurrent State-Space Model (RSSM)~\cite{hafner2019learning} to capture continuous navigational dynamics within individual VLN episodes by modeling latent state transitions.
%Our world model employs the Recurrent State-Space Model (RSSM)~\cite{hafner2019learning} to capture the navigational dynamics within individual VLN episodes by modeling latent state transitions.  
%Extending this foundation, NavMorph incorporates self-evolution during online testing, enabling agents to accumulate cross-episodic insights for comprehensive environmental understanding and effective adaptation to novel scenes. 
Specifically, the framework comprises two core networks: World-aware Navigator and Foresight Action Planner. The World-aware Navigator infers environmental dynamics from historical context and current observations, constructing robust latent state representations. The Foresight Action Planner then leverages these latent states to learn a navigation policy that outputs control signals, while decoding them into foresight visual embedding, thus providing enriched scene-aware visual information for navigation decisions.
%providing scene-aware guidance for decision-making. providing enriched scene knowledge and guidance for plausible future predictions.
To further equip the model with evolving adaptability, we introduce Contextual Evolution Memory (CEM), a simple yet effective memory mechanism seamlessly integrated into two networks of our model. CEM serves as a central enhancement to the recurrent module within RSSM, bolstering foresight planning and plausible navigational inference. 
Moreover, CEM enables efficient and dynamic updates during learning and online testing, empowering NavMorph with self-evolving abilities.

%Specifically, CEM accumulates and consolidates scene-contextual latent representations  learned from historical interactions with the environment in an unsupervised manner. At each action step within a VLN task, the agent retrieves relevant contextual features from this memory, supporting more informed decision-making and action execution across various environments.
%This innovative design endows the VLN agent with comprehensive environmental perception and efficient self-evolving ability.

Our contributions can be summarized as follows: 
\begin{itemize}[left=1em]
	\item Starting from the definition, a novel self-evolving world model is developed for VLN-CE tasks, which adaptively models latent representations of continuous environments,  enabling effective foresight-driven decision-making and adaptability in online navigation.
	%Built on the RSSM structure, we propose a self-evolving world model (NavMorph) for VLN in continuous environments, empowered with evolving environmental modeling and foresight prediction capabilities, to meet the demands for robust adaptability and effective navigation in online scenarios .
	\item We design World-aware Navigator to infer latent representations of environmental dynamics and Foresight Action Planner to optimize navigational policy via predictive modeling, integrating seamlessly with Contextual Evolution Memory to accumulate navigation insights for plausible and dynamic action planning in evolving environmental observations.
	
	%World-aware Navigator and Foresight Action Planner are incorporated to enhance environmental understanding and policy learning, while the Contextual Evolution Memory is elaborately designed to enriches contextual semantics for  enhanced generalization, bridging the performance gap between seen and unseen environments. 
	\item Extensive experiments on popular VLN-CE benchmarks demonstrate that NavMorph significantly enhances the performance of several leading models, validating its improved adaptability and generalization.
	%Our NavMorph elevates the performance of several leading VLN-CE models on two popular benchmarks. Extensive experiments on the R2R-CE and RxR-CE datasets demonstrate its enhanced adaptability and generalization in handling online navigation tasks.
	
\end{itemize}

\vspace{-3mm}
\begin{figure*}[t!]
	\centering
	\includegraphics[width=0.99\linewidth]{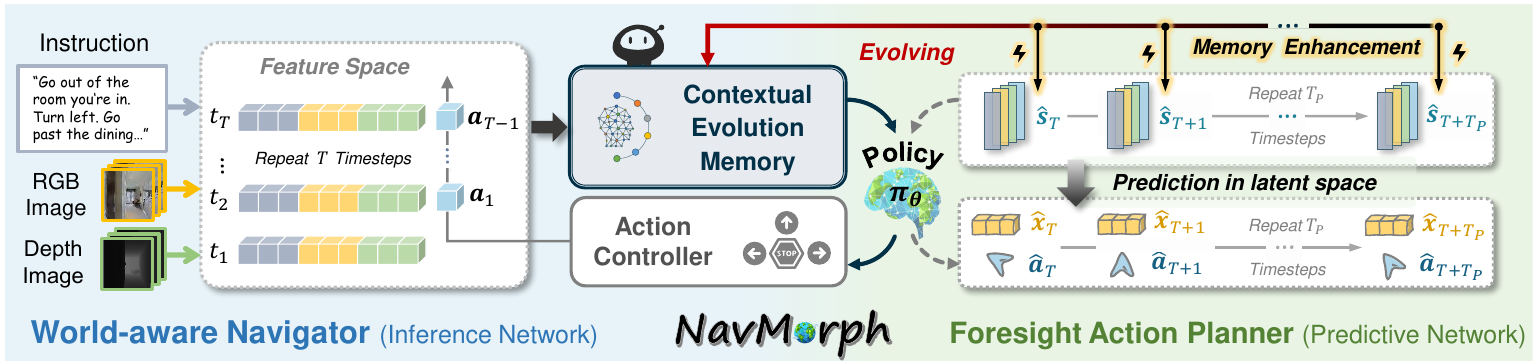}
	\vspace{-3mm}
	\caption{Overall framework of the proposed NavMorph, a self-evolving world model for VLN-CE tasks. World-aware Navigator (Inference Network) infers latent representations by leveraging current observations $\boldsymbol{o}$ and historical context from Contextual Evolution Memory. Foresight Action Planner (Predictive Network) then uses these latent representations $\boldsymbol{\hat{s}}$ to predict future visual embeddings $\boldsymbol{\hat{x}}$, enriching scene semantics for action sequence prediction $\boldsymbol{\hat{a}}$. This integrated framework enables robust navigation and effective policy learning.		
	}
	\label{fig:EWM}
	\vspace{-6mm}
\end{figure*}

\section{Related Work}\label{sec:rw}
\vspace{-2mm}
\noindent\textbf{VLN in Continuous Environments.}
Recently, Vision-and-Language Navigation (VLN) tasks~\cite{anderson2018vision,qi2020reverie,banerjee2021robotslang, an2023bevbert,qiao2023hop+,li2024panogen,he2024frequency} have garnered considerable attention among embodied AI, requiring an agent to follow human language instructions to navigate in previously unseen environments using visual observations.
To remove the unrealistic assumptions in discrete VLN settings, VLN in continuous environments (VLN-CE)~\cite{krantz2020beyond} is proposed to extend navigation to continuous spaces. 
In VLN-CE, agents can freely explore any unobstructed location in 3D environments, using low-level actions (\textit{e.g.,} MOVE-FORWARD 0.25m, TURN-LEFT/RIGHT $15^{\circ}$ and STOP) through the Habitat Simulator~\cite{savva2019habitat}. 
Early VLN-CE methods employed end-to-end trained systems to directly predict low-level actions from nature language instructions and visual observations~\cite{krantz2020beyond,irshad2022semantically}. To alleviate the reliance on costly, data-intensive pre-training, Krantz \etal~\cite{krantz2021waypoint} proposed to decouple the task into subgoal planning and low-level control by waypoint predictor.
%, where agents select from navigable sub-goals estimated by waypoint predictor and reach them through a rotate-then-forward control flow. 
Building on this framework, recent studies have explored waypoint-based approaches by using more robust visual representations pre-trained on large datasets~\cite{kamath2023new}, incorporating graph-based models~\cite{an2023bevbert}, or integrating auxiliary tasks~\cite{hong2023learning} to improve performance.
Currently, many of these models rely on panoramic RGB-D inputs~\cite{an2024etpnav,Hong2022BridgingTG,anderson2021sim,Krantz2022Sim2SimTF}, but their high cost, bulkiness, and computational overhead hinder deployment on physical robots. Therefore, the monocular configuration—processing single RGB-D frames sequentially—has emerged as a practical alternative, positioning it as a favorable approach for real-world applications in state-of-the-art methods~\cite{wang2024sim,zhang2024navid}.

\vspace{-1mm}
\noindent\textbf{World Models.}
World models have emerged as a promising framework for data-efficient learning in simulated environments~\cite{hu2022model,wu2023daydreamer,min2024driveworld,wang2024driving,poudel2024recore} and video games~\cite{hafner2023mastering,zhang2024storm}. These generative models embed observations into latent states, predict future states conditioned on actions, and decode latent predictions back into observation space~\cite{lecun2022path}. 
As an effective framework for latent dynamics learning, RSSM consists of an Inference Network for estimating latent states and a Predictive Network for forecasting future transitions. Ha \etal~\cite{ha2018recurrent} introduced a prototypical two-part model combining RSSM with a variational autoencoder.
%While RSSM provides an effective framework for latent dynamics learning, consisting of an Inference Network for estimating latent states and a Predictive Network for forecasting future transitions, Ha \etal~\cite{ha2018recurrent} introduced a prototypical two-part model, where a variational autoencoder reconstructs observations, and RSSM predicts future states.% based on the current state and action.
%While RSSM offers an effective solution for latent dynamics learning consisting of inference and predictive (generative) network, Ha \etal~\cite{ha2018recurrent} defined the prototypical two-part model where a variational auto-encoder reconstructs observations, and an RSSM predicts future latent states conditioned on the current state and action. 
Subsequent works like PlaNet~\cite{hafner2019learning} and the Dreamer series~\cite{hafner2020dream,okada2022dreamingv2,hafner2023mastering} refined this approach, optimizing policies over predicted latent states. %DriveWorld~\cite{min2024driveworld} further extended this concept to autonomous driving by integrating multi-camera inputs for enhanced spatiotemporal representations.
World models enable agents to model environmental dynamics and predict future states without decoding observations, which is crucial for reasoning and planning in VLN tasks.
%World models target at establishing dynamic models of environments, allowing agents to predict future states without decoding back to the observation space. This aspect is of paramount importance in VLN tasks, where precise predictions are essential for agent reasoning and planning.
However, their application in VLN remains underexplored. Pathdreamer~\cite{koh2021pathdreamer} introduced a visual world model to generate future observations at discrete viewpoints based on prior seen views, using them as augmented data for training. In continuous settings, DREAMWALKER~\cite{Wang2023DREAMWALKERMP} adopted a heuristic search-based world model to generate panoramic images of navigable candidates.  
Recently, Navigation World Models~\cite{Bar2024NavigationWM} introduced a controllable video generation model that utilizes learned visual priors to imagine trajectories in unfamiliar environments.
Despite notable gains, existing navigational world models encounter challenges in VLN tasks, as they (1) rely on discrete state dynamics, limiting their ability to model continuous action-state transitions, (2) use static pretrained models, restricting adaptability to novel and dynamic environments, and (3) employ pixel-level future prediction that may incur high computational costs.
To address these limitations, our proposed world model that models continuous action-state transitions through latent representation learning, adapts dynamically via a self-evolving mechanism, and reconstructs visual embeddings and perform future prediction at feature level to enhance efficiency.
%Despite notable gains, these methods~\cite{paster2021blast,sun2024learning,deng2022dreamerpro} face challenges as their pixel-wise training objectives with image reconstruction may prioritize local RGB values over essential semantic information and incur high computational costs in complex environments.
%Different from prior navigational world models rely on pixel-level generation, our world model performs foresight action planning by reconstructing visual embeddings at the feature level, thereby alleviating the computational burden while learing latent dynamics efficiently. 
%While most current methods rely on static pretrained models, our method employs an evolving mechanism that dynamically updates latent representations in an online manner, enhancing robustness and adaptability to dynamic environments.
\vspace{-3mm}
\vspace{-2.5mm}
\section{Method}\label{sec:method}
\vspace{-2mm}
In this section, we propose a self-evolving world model tailored for VLN-CE~\cite{krantz2020beyond}, a practical setting where the agent navigates a 3D mesh environment through low-level, parameterized actions. 
Our model aims to learn and adapt to the spatio-temporal dynamics of continuous environments within a structured latent space, facilitating effective reasoning and action planning during online testing. %Unlike conventional methods~\cite{koh2021pathdreamer,Wang2023DREAMWALKERMP}, our evolving model constantly refines the latent representation to capture the underlying environmental dynamics, allowing the agent to anticipate future states and make context-aware decisions. 
%We explore the task of instruction-following navigation in indoor environments, where an agent must interpret and follow a natural language instruction to reach the specified target.
%In this section, we focus on Vision-and-Language Navigation in Continuous Environments (VLN-CE)~\cite{}, a practical setting in which the agent navigates a 3D mesh representation of the environment by executing low-level, parameterized actions (\textit{e.g.,} MOVE-FORWARD 0.25m, TURN-LEFT/RIGHT $15^{\circ}$ and STOP). 
%The goal is to capture the spatio-temporal, topological and dynamic properties of continuous navigation environments, which are subsequently modeled within a structured and compact latent space. The learned latent representation enables great foresight to capture the expected future events, enhancing the process of learning from past experiences and current observations.
%Our focus lies on Vision-and-Language Navigation in Continuous Environments (VLN-CE)~\cite{krantz2020beyond}, a practical setting in which the agent navigates a 3D mesh representation of the environment by executing low-level, parameterized actions. 

\noindent\textbf{Task Definition.}
VLN-CE~\cite{krantz2020beyond} utilizes the Habitat Simulator~\cite{savva2019habitat} to render observations of environments derived from the Matterport3D dataset~\cite{chang2017matterport3d}, providing the agent with RGB-D sensory inputs. 
Each episode, referring to a complete navigation task, begins with the VLN agent receiving an  instruction alongside an initial visual observation, and concludes when the agent either selects the `stop' action or reaches the maximum step limit.
At each timestep $t$, the agent predicts waypoints~\cite{krantz2021waypoint,Hong2022BridgingTG} from a predefined set of angles and distances, which are subsequently translated into low-level control actions. This navigation progress unfolds over time steps $t\in \mathbb{N}$, with high-dimensional RGB-D images $\boldsymbol{o}_t \in \mathbb{R}^{N_o \times h \times w \times c}$ serving as input observations to infer navigational actions $\boldsymbol{a}_t =  \pi(\boldsymbol{a}_t|\boldsymbol{o}_{1:t},\boldsymbol{a}_{1:t-1})$ under a learnable policy $\pi$.  
The sequence $\boldsymbol{o}_{1:t}$ denotes all observations up to $t$, while $\boldsymbol{a}_{1:t-1}$ denotes all navigation actions up to $t-1$.

\noindent\textbf{Framework Overview.}
To enhance the agent's ability to understand and interact with environments for VLN-CE tasks, as illustrated in Figure~\ref{fig:EWM}, our proposed world model consists of two primary components: (1) World-Aware Navigator (Inference Network), which continuously constructs environmental representations and infers state transitions in a latent space, and (2) Foresight Action Planner (Predictive Network), which predicts future states to facilitate effective policy learning for strategic navigation.

\begin{figure}
	\centering
	\includegraphics[width=1\linewidth]{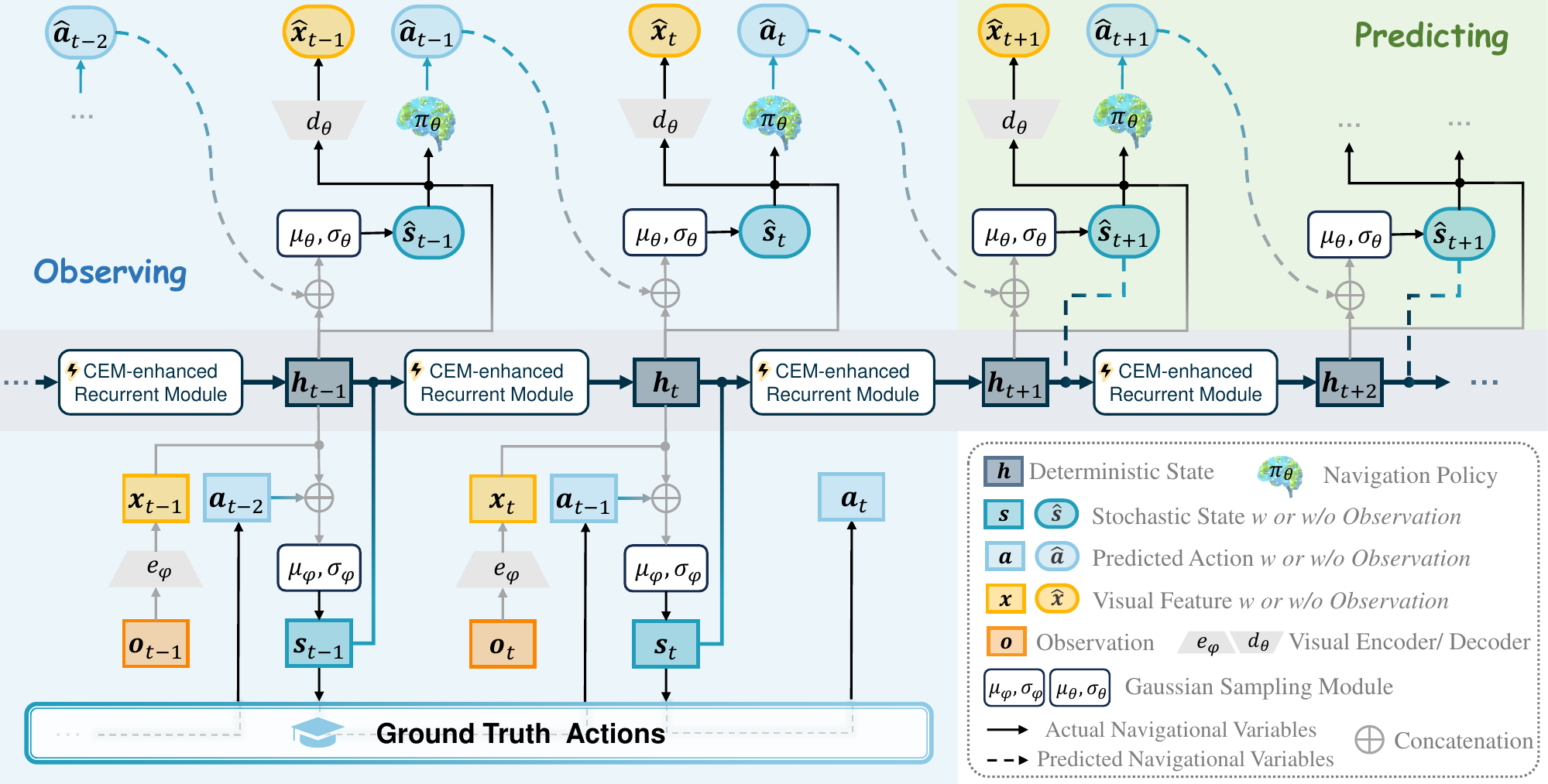}
	\vspace{-7mm}
	\caption{Detailed architecture of our world model.}
	\label{fig:RSSM}
	\vspace{-6mm}
\end{figure}

Within the World-Aware Navigator (parameterized by $\phi$), we adopt a structured latent representation following RSSM~\cite{hafner2019learning}, decomposed into two components: a deterministic history $\boldsymbol{h}_{t}$ and a stochastic state $\boldsymbol{s}_{t}$, maintaining consistent navigation progress while incorporating new observations and their potential variations. A visual encoder $e_{\phi}$ processes observations and integrates them with these latent variables $(\boldsymbol{h}_{t-1},\boldsymbol{s}_{t-1})$ to form a comprehensive state representation $\boldsymbol{h}_{t}$ for navigation decision-making.
The Predictive Network (parameterized by $\theta$) then leverages this latent representation to plan foresight actions sequentially. Specifically, it generates the predicted latent stochastic feature $\boldsymbol{\hat{s}}_t$, which encapsulates potential environmental uncertainties in future time steps. Subsequently, a visual decoder $d_{\theta}$ transforms this feature into a predicted visual embedding $\boldsymbol{\hat{x}}_t$, providing a high-level semantic representation of the anticipated navigational scene to guide the prediction of the next action $\boldsymbol{\hat{a}}_{t+1}$.
Notably, a recurrent module $f_{\phi/\theta}$ is utilized as the fundamental component of RSSM for modeling latent states, sharing parameter between both above components to ensure consistent temporal dynamics across state inference and future prediction. For notational clarity, we denote it as $f$.
An overview of detailed architecture is shown in Figure~\ref{fig:RSSM}, with components of our self-evolving world model outlined below:

\begin{itemize}
	\item \textbf{Inference Network}\vspace{-3mm}
	\begin{align*}
		\text{Visual Representation:}& \,\boldsymbol{x}_t = e_{\phi}(\boldsymbol{o}_t)\\
		\text{Initial Latent Deterministic State:} & \,\boldsymbol{h}_1 \sim \delta(\mathbf{0})\\
		\text{Recurrent Model:}& \,\boldsymbol{h}_{t} = f(\boldsymbol{h}_{t-1}, \boldsymbol{s}_{t-1})\\
		\text{Dynamic Transition Model:}& \,\boldsymbol{s}_{t} \sim q_{\phi}(\boldsymbol{s}_{t}|\boldsymbol{o}_{1:t}, \boldsymbol{a}_{1:t-1})
	\end{align*}%\vspace{-6mm}
	\item \textbf{Predictive Network}\vspace{-3mm}
	\begin{align*}
		\text{Initial Latent Stochastic State:}& \,\boldsymbol{s}_1 \sim \mathcal{N}(\boldsymbol{0,I})\\
		\text{Stochastic State Model:}& \,\boldsymbol{\hat{s}}_{t} \sim p_{\theta}(\boldsymbol{\hat{s}}_{t}|\boldsymbol{h}_{t},\boldsymbol{\hat{s}}_{t-1})\\
		\text{Visual Decoder:}& \,\boldsymbol{\hat{x}}_t \sim p_{\theta}(\boldsymbol{\hat{x}}_t|\boldsymbol{h}_{t},\boldsymbol{\hat{s}}_{t})\\
		%\text{Action Decoder:}& \,\boldsymbol{\hat{a}}_t \sim {p}_{\theta}(\boldsymbol{\hat{a}}_t|\boldsymbol{h}_{t}, \boldsymbol{\hat{s}}_{t})
		\text{Action Decoder:}& \,\boldsymbol{\hat{a}}_t \sim {p}_{\theta}(\boldsymbol{\hat{a}}_t|\boldsymbol{h}_{t}, \boldsymbol{\hat{s}}_{t})
	\end{align*}
\end{itemize}

\vspace{-4mm}
\subsection{World-Aware Navigator (Inference Network)}\label{sec:3.1}
\vspace{-1mm}
The World-Aware Navigator is a policy-based agent designed to effectively map the cross-modal input onto a sequence of actions within the VLN framework.  
%containing an RSSM to integrate past observations and contextual cues to learn a comprehensive latent representation of the environment.s~\cite{wu2023daydreamer, min2024driveworld},
Actions are pivotal for the world model to anticipate and predict future states. In a typical VLN-CE setting~\cite{koh2021pathdreamer,an2024etpnav}, the agent predicts actions by selecting a candidate sub-goal point on a constructed map, along with its corresponding coordinates. To ensure plausible predictions of NavMorph, we define the action at time step $t$ as $ \Delta position_t$, which represents the movement of ego location between the previous time step $t-1$ to current time step $t$. For consistency with operations in the latent space of our world model, a linear transformation maps the action into a $d_a$-dimensional embedding $\boldsymbol{a}_t \in \mathbb{R}^{d_a}$. 
After integrating the encoded action with the agent's current state and observation, we sample from $\left( \boldsymbol{h}_t, \boldsymbol{a}_t, \boldsymbol{x}_t \right)$ to estimate the Gaussian posterior, yielding a probabilistic representation of the current state:
\vspace{-2.5mm}
\begin{align}\label{eq:posterior}
	q(\boldsymbol{s}_t|\boldsymbol{o}_{1:t}, \boldsymbol{a}_{1:t-1}) \sim \mathcal{N}(\mu_{\phi}(&\boldsymbol{h}_t,\boldsymbol{a}_{t-1},\boldsymbol{x}_t),
	\notag
	\\ &\sigma_{\phi}(\boldsymbol{h}_t,\boldsymbol{a}_{t-1},\boldsymbol{x}_t)\boldsymbol{I}),
\end{align}
where $(\mu_{\phi},\sigma_{\phi})$ are multi-layer perceptions (MLPs) that parameterize the posterior state distribution. 
During training, $\boldsymbol{a}_{t}$ corresponds to ground truth actions.

In our model, the deterministic history should encode both temporal dynamics and accumulated information from past observations and language instructions. A common approach to achieve this is through conventional recurrent modules like RNNs, LSTMs, or transformer-based architectures, which can process current observations alongside previously obtained features to recursively integrate historical information $\boldsymbol{h}_{t-1}$ and stochastic states $\boldsymbol{s}_{t-1}$ into a full history representation $\boldsymbol{h}_t$. 
However, while effective in capturing short-term dynamics, these approaches struggle with adaptive updates during online testing due to their reliance on gradient descent and lack an explicit memory mechanism to store and utilize novel scene information.
%However, while effective at capturing short-term dynamics and providing coherent understanding immediate experiences, these approaches often struggle to adaptive updates during online testing due to the reliance on gradient descent, and lack an explicit memory mechanism to store and utilize environmental information effectively.
%often struggle to store and use semantic knowledge across various testing episodes, particularly in online navigation where agents encounter new room layouts, unobserved objects, or unexpected obstacles that deviate from training distributions.
Therefore, we introduce a simple yet effective Contextual Evolution Memory (CEM), which enhances the agent's capacity to retain and leverage historical navigation insights. CEM maintains $N_m$ scene-contextual features $\left\{ \boldsymbol{v}_m \right\}_{m=1}^{N_m}$, where each $\boldsymbol{v}_m \in \mathbb{R}^{d_v}$ captures crucial visual insights from diverse scenarios encountered.
To accommodate the dynamic nature of navigation environments, CEM continuously evolves by incorporating new scene features while selectively preserving valuable historical context. 
CEM is randomly initialized and progressively integrates informative scene-contextual features accumulated throughout learning and online testing. 
%CEM initially starts in a randomized state and progressively integrates informative scene-contextual features accumulated throughout learning and online testing.
This adaptive mechanism ensures relevance to recent experiences while preserving critical visual-semantic information from past episodes.
Specifically, this evolving CEM supports the enhanced deterministic history $\Tilde{\boldsymbol{h}}_t$ by assimilating relevant contextual information from past navigational experiences. Our CEM-enhanced recurrent module retrieves the top-$K$ scene-contextual features from CEM, using them as historical semantic insights to enrich the deterministic state:\vspace{-3mm}
\begin{align}
	\begin{split}
		\left\{ \boldsymbol{v}_k \right\}_{k=1}^K &= \operatorname{arg\,top\mathit{K}} \left( \mathcal{SIM}(\boldsymbol{h}_t, \boldsymbol{v}_m) \right), \\
		\tilde{\boldsymbol{h}}_t &= (1 - \alpha) \boldsymbol{h}_t + \alpha \sum_{k=1}^K \mu_k \boldsymbol{v}_k,
	\end{split}
\end{align}
where the relevance score $\mathcal{SIM}(\boldsymbol{h}_t, \boldsymbol{v}_m)$ is calculated using cosine similarity. The weight $\mu_{k}$ denote the normalized similarity score assigned to the $k$-th relevant retrieved features, while $\alpha$ serves as an enhancing factor.
% that controls the balance between the original state and the contribution from the retrieved memory. to the current scene

To balance the trade-off between previously accumulated experiences and newly observed information, CEM is efficiently evolved by refining the retrieved scene-contextual features through forward updates, rather than using time-consuming gradient-based backpropagation. This evolving process is governed by a factor $\beta$ and the enriched deterministic state is then used for future predictions:\vspace{-2mm}
\begin{align}\label{eq:memory}
	\boldsymbol{v}_k \leftarrow (1- \beta)\boldsymbol{v}_{k}+\beta\boldsymbol{h}_{t}, \;
	\boldsymbol{h}_t \leftarrow \tilde{\boldsymbol{h}}_t.
\end{align}

\vspace{-4mm}
\subsection{Foresight Action Planner (Predictive Network) }\label{sec:3.2}
\vspace{-1mm}
Empowered by the world model, the agent acquires the ability to anticipate environmental changes resulting from its actions over an extended future period. Based on the state representations learned by the World-Aware Navigator, the Foresight Action Planner is designed to execute future planning by imagining potential scenarios in the latent space. 
In contrast to future predictions with observed images, we rely solely on the enhanced history states $\boldsymbol{h}_{1:t}$ provided by the CEM-enhanced recurrent module, and the predicted states $\boldsymbol{\hat{s}}_{1:t}$. Similar to inference network, $\boldsymbol{\hat{s}}_t$ is sampled from $(\boldsymbol{h}_t, \boldsymbol{\hat{a}}_t)$ with the Gaussian prior distribution, defined as:\vspace{-1mm}
\begin{align}\label{eq:prior}
	p_{\theta}(\boldsymbol{\hat{s}}_{t}|\boldsymbol{h}_{t},\boldsymbol{\hat{s}}_{t-1}) \sim \mathcal{N}(\mu_{\theta}(\boldsymbol{h}_{t},\boldsymbol{\hat{a}}_{t-1}), \sigma_{\theta}(\boldsymbol{h}_{t},\boldsymbol{\hat{a}}_{t-1})\boldsymbol{I}),
\end{align}
where $(\mu_{\theta},\sigma_{\theta})$ are MLPs that parameterize the prior state distribution. Since the prior does not have access to the ground truth actions, the predicted action $\boldsymbol{\hat{a}}_{t-1}$ is estimated with the learned policy ${\pi}_{\theta}( \boldsymbol{h}_{t-1}, \boldsymbol{\hat{s}}_{t-1})$.
%with $T$ denoting the total number of steps taken by the agent with observing inputs. 
Building upon this, the Foresight Action Planner extends the prediction of future states over $T_p$ time steps without access to corresponding observations, leveraging the learned latent state and previously inferred actions over previous $T$ timesteps (\textit{i.e.}, observation window). In specific, the planner applies the learned policy to estimate actions $\boldsymbol{\hat{a}}_{T+j}=\pi_{\theta}(\boldsymbol{h}_{T+j},\boldsymbol{\hat{s}}_{T+j})$, predicts the next deterministic state $\boldsymbol{h}_{T+j+1}=f_{\theta}(\boldsymbol{h}_{T+j},\boldsymbol{\hat{s}}_{T+j})$ and samples from the prior distribution: $\boldsymbol{\hat{s}}_{T+j+1} \sim \mathcal{N}(\mu_{\theta}(\boldsymbol{h}_{T+j+1},\boldsymbol{\hat{a}}_{T+j}), \sigma_{\theta}(\boldsymbol{h}_{T+j+1},\boldsymbol{\hat{a}}_{T+j})\boldsymbol{I})$, $j\in \left [ 1,T_p \right ] $. 

Aiming for a comprehensive understanding of environmental dynamics to support action planning, we introduce a reconstruction task that guides foresight action planner to simulate potential future scenarios internally while learning useful latent representations.
To achieve this, we reconstruct visual embeddings $\boldsymbol{x}_t \in \mathbb{R}^{d_x}$ derived from input observations $\boldsymbol{o}_t$ through a visual decoder $d_{\theta}$, rather than direct pixel-level image reconstruction. 
While pixel-level generation methods~\cite{li2023improving,koh2021pathdreamer, Wang2023DREAMWALKERMP,Bar2024NavigationWM} are common for future observation prediction, they incur high computational costs and training complexity, while their reliance on pretrained generative models may introduce biases and hinder fair comparisons.
%While pixel-level generation methods~\cite{li2023improving,koh2021pathdreamer, Wang2023DREAMWALKERMP,Bar2024NavigationWM} are common for future observation imagination, they incur increased computational costs and training complexity due to high dimensionality of image data. Moreover, their dependence on pretrained generative models may introduce external biases and complicate fair comparisons across approaches. 
This reconstruction objective serves to regularize and enhance the model’s future state predictions, with details provided in \S~\ref{sec:3.3}. Additionally, the designed CEM enriches latent representations with contextually relevant historical information, leading to more informed reconstructions and improved planning.

\noindent\textbf{Discussion.} 
%The above analysis highlights the crucial role of CEM in our proposed world model, serving as a core component within RSSM to facilitate plausible foresight planning and enable efficient self-evolution.
%Instead of designing a complex evolution mechanism, we adopt a simple and direct approach that dynamically integrates scene-contextual information during learning and online testing.
%Given the limited exploration of adaptive evolution in existing world models, we aim to demonstrate the potential of incorporating evolution strategies into world models in VLN-CE tasks. 
The above analysis highlights the crucial role of CEM in our proposed world model, serving as a core component within RSSM to facilitate plausible foresight planning and enable efficient self-evolution. Given the limited exploration of dynamic evolution within existing world models, this paper aims to elucidate the potential of integrating evolutionary strategies into VLN world models. To this end, instead of designing a more complex evolution mechanism, we design a simple, direct, and effective approach that dynamically integrates scene contextual information during learning and online testing.
Refer to Sec~\ref{sec:4.3} and \textbf{Supplementary Material} for further discussion. 

\vspace{-1mm}
\subsection{Pre-training Objective}\label{sec:3.3}
\vspace{-1mm}
%\noindent\textbf{World Model Training.} To enable an in-depth understanding of latent dynamics and the underlying data distribution, omit the conditioning set of the posterior distribution $q(\boldsymbol{s}_t|\boldsymbol{o}_{1:t},\boldsymbol{a}_{1:t-1})$ and \S~\ref{sec:ELBO}
Our evolving world model is trained by maximizing a variational lower bound on the data log-likelihood. This training process involves minimizing losses related to visual reconstruction and action prediction for both past and future states, as well as the divergence between the posterior and prior state distributions. Specifically, the model observes inputs over $T$ timesteps and subsequently forecasts visual embeddings and actions for an additional $T_p$ timesteps. The loss function $\mathcal{L}_W$ is defined as follows:\vspace{-2mm}
\begin{equation*}
		\mathcal{L}_{W} = \sum_{t=1}^{T} \mathbb{E} \left[ \underbrace{-\log p(\boldsymbol{x}_{t} | \boldsymbol{h}_{t}, \boldsymbol{s}_{t})}_{\text{Reconstruction Loss $\ell_{re}$}} \underbrace{-\log p(\boldsymbol{a}_{t} | \boldsymbol{h}_{t}, \boldsymbol{s}_{t})}_{\ell_{ac}} \right] 
\end{equation*}
\vspace{-5mm}
\begin{align}
	\begin{split}
		&+ \sum_{j=1}^{T_p} \mathbb{E}\left[ \underbrace{-\log p(\boldsymbol{x}_{T+j} | \boldsymbol{h}_{T}, \boldsymbol{s}_{T})}_{\ell_{re}} \underbrace{-\log p(\boldsymbol{a}_{T+j} | \boldsymbol{h}_{T}, \boldsymbol{s}_{T})}_{\text{Action Prediction Loss $\ell_{ac}$}} \right] \\
		&+ \sum_{t=1}^{T} \mathbb{E} \left[ \gamma \cdot \underbrace{D_{\mathrm{KL}} \left( q(\boldsymbol{s}_{t} | \boldsymbol{o}_{1:t}, \boldsymbol{a}_{1:t-1}) \, \| \, p(\boldsymbol{s}_{t} | \boldsymbol{h}_{t-1}, \boldsymbol{s}_{t-1}) \right)}_{\text{Posterior and Prior Matching Loss $\ell_{kl}$}} \right],
	\end{split}
\end{align}
where $\gamma$ is a weighting coefficient. For simplicity of notation, we use a unified notation ($\boldsymbol{s}_t, \boldsymbol{a}_t, \boldsymbol{x}_t$) to represent variables across both the observation and prediction phases, omitting the hat notation ($\hat{\boldsymbol{s}}_t, \hat{\boldsymbol{a}}_t, \hat{\boldsymbol{x}}_t$). Please refer to \textbf{Supplementary Material} for the detailed derivation of $\mathcal{L}_W$.

\begin{comment}
	\begin{align}
		\begin{split}
			&\mathcal{L}_{W} = \sum_{t=1}^{T} \mathbb{E} \left[ \underbrace{-\log p(\boldsymbol{x}_{t} | \boldsymbol{h}_{t}, \boldsymbol{s}_{t})}_{\text{Reconstruction Loss $\ell_{re}$}} \underbrace{-\log p(\boldsymbol{a}_{t} | \boldsymbol{h}_{t}, \boldsymbol{s}_{t})}_{\ell_{ac}} \right] \\
			&+ \sum_{j=1}^{T_p} \mathbb{E}\left[ \underbrace{-\log p(\boldsymbol{x}_{T+j} | \boldsymbol{h}_{T}, \boldsymbol{s}_{T})}_{\ell_{re}} \underbrace{-\log p(\boldsymbol{a}_{T+j} | \boldsymbol{h}_{T}, \boldsymbol{s}_{T})}_{\text{Action Prediction Loss $\ell_{ac}$}} \right] \\
			&+ \sum_{t=1}^{T} \mathbb{E} \left[ \gamma \cdot \underbrace{D_{\mathrm{KL}} \left( q(\boldsymbol{s}_{t} | \boldsymbol{o}_{1:t}, \boldsymbol{a}_{1:t-1}) \, \| \, p(\boldsymbol{s}_{t} | \boldsymbol{h}_{t-1}, \boldsymbol{s}_{t-1}) \right)}_{\text{Posterior and Prior Matching Loss $\ell_{kl}$}} \right],
		\end{split}
	\end{align}
\end{comment}
In practice, we implement the log-likelihood term with two components: a visual embedding reconstruction loss $\ell_{re}$ and an action prediction loss $\ell_{ac}$,  optimized using stochastic gradient ascent.  
While point-wise difference metrics (\textit{e.g.}, cross-entropy) provide a direct measure of prediction accuracy, they  may not adequately capture the temporal nature of navigation trajectories and potential misalignments between predicted and actual routes. To mitigate this, we incorporate a regularization term based on Normalized Dynamic Time Warping (NDTW)~\cite{ilharco2019general}, which constrains the predicted visual embeddings and actions. This regularization is specifically formulated as follows:\vspace{-1mm}
\begin{align}
	\ell_{\star} = 1-  \frac{1}{T_p}\sum_{t=1}^{T} \sum_{j=1}^{T_p}\text{NDTW} (\star_{1:t+j},\hat{\star}_{1:t+j}),
\end{align}
where $\star_{1:t+j}$ and $\hat{\star}_{1:t+j}$ denote the ground-truth and predicted sequences, respectively.
This NDTW-based regularization is applied to both $\ell_{re}$ and $\ell_{ac}$, encouraging temporal consistency and alignment between predicted and actual navigation trajectories. Additionally, to ensure precise action predictions, we incorporate an L2 loss within $\ell_{ac}$, focusing on accurate estimation of $\boldsymbol{a}_t = \Delta position_t$ and leveraging supervision from ground truth actions.

In addition to ensuring foresight latent states $\boldsymbol{\hat{s}}_{t}$ align with observed data, a Kullback-Leibler (KL) divergence-based loss is employed to align the prior distribution with the posterior.
Since both the approximate $q(\boldsymbol{s}_{t} | \boldsymbol{o}_{1:t}, \boldsymbol{a}_{1:t-1})$ and the prior $p(\boldsymbol{s}_{t} | \boldsymbol{h}_{t-1}, \boldsymbol{s}_{t-1}) $ are modeled as Gaussian distributions $\left(\right.$Eq.~\eqref{eq:posterior} and Eq.~\eqref{eq:prior}$\left.\right)$, the posterior and prior matching loss can be computed in closed form. Upon convergence, the KL divergence quantifies the information disparity between the prior and posterior distributions, ensuring the prior models state-action dynamics observed in actual navigation.
Finally, the world model loss $\mathcal{L}_W$ is combined with the imitation learning objective $\mathcal{L}_{IL}$~\cite{an2024etpnav}, to form the complete loss function: $\mathcal{L} = \mathcal{L}_{W} + \mathcal{L}_{IL}$. Note that $\mathcal{L}_{IL} = - \sum_{t=1}^{T}\log p(\boldsymbol{a}^*|\mathcal{I}, \mathcal{O})$, and $\boldsymbol{a}^*$ represents the teacher action node of each step, determined by an interactive demonstrator similar to DAgger algorithm~\cite{ross2011reduction}.	

\vspace{-2mm}
\subsection{Working Modes}\label{sec:3.4}\vspace{-2mm}
Our world model operates in two distinct modes during the training and testing phases.

\noindent\textbf{During training}, the model focuses on learning the underlying latent dynamics of environments through future state prediction. It processes sequences of observations and corresponding actions, optimizing loss functions delineated in \S~\ref{sec:3.3} to update the model via gradient descent.%, while CEM continuously performs self-evolution, refining latent representations to support robust navigational policy learning.

\noindent\textbf{During testing,} our self-evolving world model continually adapts to changing environments via the evolving mechanism. 
The recurrent module updates CEM in an online manner during testing, similar to the training phase as described in Eq.~\eqref{eq:memory}. This adaptive ability allows the model to retain insightful information from recent observations while adjusting to testing scenes, thereby maintaining a contextually enriched and up-to-date comprehension of its surroundings.

%To imbue the model with temporal modeling ability, we introduce two latent variables in alignment with the structure of RSSM: the deterministic history $\boldsymbol{h}_{t}$ and the stochastic state $\boldsymbol{s}_t$. Specifically, the World-Aware Navigator (parameterized by $\phi$) processes observational inputs through a visual encoder $e_{\phi}$, enabling the agent to build a robust latent representation from accumulated past history states $\boldsymbol{h}_{t-1}$ and latent stochastic feature $\boldsymbol{s}_{t-1}$, accommodating the unpredictability of environmental dynamics while preserving information continuity.
%Specifically, the World-Aware Navigator (parameterized by $\phi$) processes observational inputs through a visual encoder $e_{\phi}$, and then build a robust latent representation of continuous environmental dynamics. Based on RSSM structure, the latent state is decomposed into two components: a deterministic history $\boldsymbol{h}_{t}$ and a stochastic state $\boldsymbol{s}_t$, maintaining consistent navigation progress through $\boldsymbol{h}_{t-1}$ while incorporating new observations and their potential variations through $\boldsymbol{s}_{t-1}$.

\begin{table*}[thpb] \centering
	\caption{Experimental results on R2R-CE dataset. Results better than base model are shown in \textcolor{blue}{blue}. Best results for the panoramic and monocular settings are each highlighted in \textbf{bold}. * indicates experimental results that we have reproduced in this work.}
	\vspace{-3mm}
	\label{tab:r2r-ce}
	\resizebox{0.98\textwidth}{!}{
		%\Huge
		\begin{threeparttable}
		\begin{tabular}{@{}c|l|ccccccccccccccc@{}}
			\toprule
			& \multicolumn{1}{c|}{} & \multicolumn{5}{c|}{\textbf{Val Seen}} & \multicolumn{5}{c|}{\textbf{Val Unseen}} & \multicolumn{5}{c}{\textbf{Test Unseen}} \\  
			\multirow{-2}{*}{\textbf{Camera}} & \multicolumn{1}{c|}{\multirow{-2}{*}{\textbf{Methods}}} & \cellcolor{red!25}TL $\downarrow$ & \cellcolor{red!25}NE $\downarrow$ & \cellcolor{gray!25}OSR & \cellcolor{gray!25}SR & \cellcolor{gray!25}SPL & \cellcolor{red!25}TL $\downarrow$ & \cellcolor{red!25}NE $\downarrow$ & \cellcolor{gray!25}OSR & \cellcolor{gray!25}SR & \cellcolor{gray!25}SPL & \cellcolor{red!25}TL $\downarrow$ & \cellcolor{red!25}NE $\downarrow$ & \cellcolor{gray!25}OSR & \cellcolor{gray!25}SR & \cellcolor{gray!25}SPL \\ \midrule
			& LAW~\cite{Raychaudhuri2021LanguageAlignedW} {\scriptsize {\color{Gray} [EMNLP21]}}& \textbf{9.34} & 6.35 & 49 & 40 & \multicolumn{1}{c|}{37} & \textbf{8.89} & 6.83 & 44 & 35 & \multicolumn{1}{c|}{31} & \textbf{9.67} & 7.69 & 28 & 38 & 25 \\
			\multicolumn{1}{c|}{} & CM$^2$~\cite{Georgakis2022CrossmodalML} {\scriptsize {\color{Gray} [CVPR22]}} & 12.05 & 6.10 & 50.7 & 42.9 & \multicolumn{1}{c|}{34.8} & 11.54 & 7.02 & 41.5 & 34.3 & \multicolumn{1}{c|}{27.6} & 13.90 & 7.70 & 39 & 31 & 24 \\
			\multicolumn{1}{c|}{} & WS-MGMap~\cite{Chen2022WeaklySupervisedMM} {\scriptsize {\color{Gray} [NeurIPS22]}}& 10.12 & 5.65 & 51.7 & 46.9 & \multicolumn{1}{c|}{\textbf{43.4}} & 10.00 & 6.28 & 47.6 & 38.9 & \multicolumn{1}{c|}{34.3} & 12.30 & 7.11 & 45 & 35 & 28 \\
			\multicolumn{1}{c|}{} & NaVid~\cite{zhang2024navid} {\scriptsize {\color{Gray} [RSS24]}}& - & - & - & - & \multicolumn{1}{c|}{-} & - & \textbf{5.47} & 49.1 & 37.4 & \multicolumn{1}{c|}{\textbf{35.9}} & - & - & - & - & - \\
			\multicolumn{1}{c|}{} & ETPNav/p~\cite{wang2024sim} {\scriptsize {\color{Gray} [CoRL24]}} & - & - & - & - & \multicolumn{1}{c|}{-} & - & 6.81 & 42.4 & 32.9 & \multicolumn{1}{c|}{23.1} & - & - & - & - & - \\ \cmidrule(l){2-17} 
			\multicolumn{1}{c|}{} & VLN-3DFF~\cite{wang2024sim} {\scriptsize {\color{Gray} [CoRL24]}}& - & - & - & - & \multicolumn{1}{c|}{-} & - & 5.95 & 55.8 & 44.9 & \multicolumn{1}{c|}{30.4} & - & 6.24 & 54.4 & 43.7 & 28.9 \\
			\multicolumn{1}{c|}{} & \cellcolor{gray!10}VLN-3DFF* & \cellcolor{gray!10}22.90 & \cellcolor{gray!10}4.92 & \cellcolor{gray!10}62.1 & \cellcolor{gray!10}52.7 & \multicolumn{1}{c|}{\cellcolor{gray!10}36.7} & \cellcolor{gray!10}26.16 & \cellcolor{gray!10}6.05 & \cellcolor{gray!10}54.9 & \cellcolor{gray!10}43.8 & \multicolumn{1}{c|}{\cellcolor{gray!10}29.4} & \cellcolor{gray!10}26.02 & \cellcolor{gray!10}6.22 & \cellcolor{gray!10}\textbf{54.7} & \cellcolor{gray!10}43.8 & \cellcolor{gray!10}28.6 \\
			\multicolumn{1}{c|}{\multirow{-8}{*}{\textbf{Monocular}}} & \cellcolor{gray!25}\textbf{NavMorph} & \cellcolor{gray!25}\textcolor{blue}{20.03} & \cellcolor{gray!25}{\textcolor{blue}{\textbf{4.58}}} & \cellcolor{gray!25}{\textcolor{blue}{\textbf{62.7}}} & \cellcolor{gray!25}{\textcolor{blue}{\textbf{55.8}}} & \multicolumn{1}{c|}{\cellcolor{gray!25}\textcolor{blue}{38.9}} & \cellcolor{gray!25}\textcolor{blue}{22.54} & \cellcolor{gray!25}\textcolor{blue}{5.75} & \cellcolor{gray!25}{\textcolor{blue}{\textbf{56.9}}} & \cellcolor{gray!25}{\textcolor{blue}{\textbf{47.9}}} & \multicolumn{1}{c|}{\cellcolor{gray!25}\textcolor{blue}{{33.2}}} & \cellcolor{gray!25}\textcolor{blue}{24.75} & \cellcolor{gray!25}{\textcolor{blue}{\textbf{6.01}}} & \cellcolor{gray!25}54.5 & \cellcolor{gray!25}{\textcolor{blue}{\textbf{45.7}}} & \cellcolor{gray!25}{\textcolor{blue}{\textbf{30.2}}} \\  \midrule[0.7pt]
			\multicolumn{1}{c|}{} 
			& Seq2Seq~\cite{anderson2018vision} {\scriptsize {\color{Gray} [CVPR18]} } & \textbf{9.26} & 7.12 & 46 & 37 & \multicolumn{1}{c|}{35} & \textbf{8.64} & 7.37 & 40 & 32 & \multicolumn{1}{c|}{30} & \textbf{8.85} & 7.91 & 36 & 28 & 25 \\
			& Sim2Sim~\cite{Krantz2022Sim2SimTF} {\scriptsize {\color{Gray} [ECCV22]} }& 11.18 & 4.67 & 61 & 52 & \multicolumn{1}{c|}{44} & 10.69 & 6.07 & 52 & 43 & \multicolumn{1}{c|}{36} & 11.43 & 6.17 & 52 & 44 & 37 \\
			& CWP-CMA~\cite{Hong2022BridgingTG} {\scriptsize {\color{Gray} [CVPR22]}}& 11.47 & 5.20 & 61 & 51 & \multicolumn{1}{c|}{45} & 10.90 & 6.20 & 52 & 41 & \multicolumn{1}{c|}{36} & 11.85 & 6.30 & 49 & 38 & 33 \\
			& CWP-BERT~\cite{Hong2022BridgingTG}{\scriptsize {\color{Gray} [CVPR22]}}& 12.50 & 5.02 & 59 & 50 & \multicolumn{1}{c|}{44} & 12.23 & 5.74 & 53 & 44 & \multicolumn{1}{c|}{39} & 13.51 & 5.89 & 51 & 42 & 36 \\
			& DREAMW~\cite{Wang2023DREAMWALKERMP} {\scriptsize {\color{Gray} [ICCV23]}}& 11.60 & 4.09 & 59 & 66 & \multicolumn{1}{c|}{48} & 11.30 & 5.53 & 49 & {59} & \multicolumn{1}{c|}{44} & 11.80 & 5.48 & 49 & {57} & 44 \\
			& GridMM~\cite{wang2023gridmm} {\scriptsize {\color{Gray} [ICCV23]}}& 12.69 & 4.21 & 69 & 59 & \multicolumn{1}{c|}{51} & 13.36 & 5.11 & 61 & 49 & \multicolumn{1}{c|}{41} & 13.31 & 5.64 & 56 & 46 & 39 \\
			%& BEVBert~\cite{an2023bevbert} {\scriptsize {\color{Gray} [ICCV23]}}& 13.98 & 3.77 & 73 & 68 & \multicolumn{1}{c|}{60} & 13.27 & 4.57 & 67 & 59 & \multicolumn{1}{c|}{50} & 15.31 & 4.70 & 67 & 59 & 50 \\ 
			& FSTTA~\cite{gao2024fast} {\scriptsize {\color{Gray} [ICML24]}}& 12.39 & 4.25 & 69 & 58 & \multicolumn{1}{c|}{50} & 11.58 & 5.27 & 58 & 48 & \multicolumn{1}{c|}{42} & 13.17 & 5.84 & 55 & 46 & 38 \\ \cmidrule(l){2-17} 
			& ETPNav~\cite{an2024etpnav} {\scriptsize {\color{Gray} [TPAMI24]}}& 11.78 & 3.95 & 72 & 66 & \multicolumn{1}{c|}{59} & 11.99 & 4.71 & 65 & 57 & \multicolumn{1}{c|}{49} & 12.87 & 5.12 & 63 & 55 & 48 \\ 
			& \cellcolor{gray!10}ETPNav* & \multicolumn{1}{c}{\cellcolor{gray!10}11.35} & \multicolumn{1}{c}{\cellcolor{gray!10}3.93} & \multicolumn{1}{c}{\cellcolor{gray!10}72} & \multicolumn{1}{c}{\cellcolor{gray!10}66} &\multicolumn{1}{c|}{\cellcolor{gray!10}59} & \multicolumn{1}{c}{\cellcolor{gray!10}11.40} & \multicolumn{1}{c}{\cellcolor{gray!10}4.69} & \multicolumn{1}{c}{\cellcolor{gray!10}64} & \multicolumn{1}{c}{\cellcolor{gray!10}57} & \multicolumn{1}{c|}{\cellcolor{gray!10}49} & \cellcolor{gray!10}12.72 & \cellcolor{gray!10}5.10 & \cellcolor{gray!10}63 & \cellcolor{gray!10}55 & \cellcolor{gray!10}48 \\
			& \cellcolor{gray!25}\textbf{NavMorph} & \multicolumn{1}{c}{\cellcolor{gray!25}11.43} &  \multicolumn{1}{c}{\cellcolor{gray!25}{\textcolor{blue}{{3.86}}}} &  \multicolumn{1}{c}{\cellcolor{gray!25}{\textcolor{blue}{73}}} &  \multicolumn{1}{c}{\cellcolor{gray!25}{\textcolor{blue}{67}}} &  \multicolumn{1}{c|}{\cellcolor{gray!25}{\textcolor{blue}{60}}} & \multicolumn{1}{c}{\cellcolor{gray!25}11.55} & \multicolumn{1}{c}{\cellcolor{gray!25}{\textcolor{blue}{4.62}}} & \multicolumn{1}{c}{\cellcolor{gray!25}{\textcolor{blue}{66}}} & \multicolumn{1}{c}{\cellcolor{gray!25}{\textcolor{blue}{59}}} & \multicolumn{1}{c|}{\cellcolor{gray!25}{\textcolor{blue}{50}}} & \cellcolor{gray!25}12.88 & \cellcolor{gray!25}{\textcolor{blue}{4.91}} & \cellcolor{gray!25}{\textcolor{blue}{64}} & \cellcolor{gray!25}{\textcolor{blue}{57}} & \cellcolor{gray!25}{\textcolor{blue}{49}}\\ \cmidrule(l){2-17} 
			& HNR~\cite{wang2024lookahead} {\scriptsize {\color{Gray} [CVPR24]}} & 11.79 & 3.67 & 76 & 69 & \multicolumn{1}{c|}{61} & 12.64 & 4.42 & 67 & 61 & \multicolumn{1}{c|}{51} & 13.03 & 4.81 & 67 & 58 & 50 \\ 
			& \cellcolor{gray!10}HNR* & \multicolumn{1}{c}{\cellcolor{gray!10}11.84} & \multicolumn{1}{c}{\cellcolor{gray!10}3.73} & \multicolumn{1}{c}{\cellcolor{gray!10}76} & \multicolumn{1}{c}{\cellcolor{gray!10}69} &\multicolumn{1}{c|}{\cellcolor{gray!10}61} & \multicolumn{1}{c}{\cellcolor{gray!10}12.76} & \multicolumn{1}{c}{\cellcolor{gray!10}4.57} & \multicolumn{1}{c}{\cellcolor{gray!10}67} & \multicolumn{1}{c}{\cellcolor{gray!10}61} & \multicolumn{1}{c|}{\cellcolor{gray!10}51} & \cellcolor{gray!10}12.92 & \cellcolor{gray!10}4.85 & \cellcolor{gray!10}67 & \cellcolor{gray!10}58 & \cellcolor{gray!10}50 \\
			\multirow{-14}{*}{\textbf{Panoramic}} & \cellcolor{gray!25}\textbf{NavMorph} & \multicolumn{1}{c}{\cellcolor{gray!25}\textcolor{blue}{11.76}} &  \multicolumn{1}{c}{\cellcolor{gray!25}\textbf{\textcolor{blue}{3.66}}} &  \multicolumn{1}{c}{\cellcolor{gray!25}\textbf{\textcolor{blue}{78}}} &  \multicolumn{1}{c}{\cellcolor{gray!25}\textbf{\textcolor{blue}{70}}} &  \multicolumn{1}{c|}{\cellcolor{gray!25}\textbf{\textcolor{blue}{62}}} & \multicolumn{1}{c}{\cellcolor{gray!25}\textcolor{blue}{12.68}} & \multicolumn{1}{c}{\cellcolor{gray!25}\textbf{\textcolor{blue}{4.37}}} & \multicolumn{1}{c}{\cellcolor{gray!25}\textbf{\textcolor{blue}{68}}} & \multicolumn{1}{c}{\cellcolor{gray!25}\textbf{\textcolor{blue}{64}}} & \multicolumn{1}{c|}{\cellcolor{gray!25}\textbf{\textcolor{blue}{53}}} & \cellcolor{gray!25}\textcolor{blue}{12.69} & \cellcolor{gray!25}\textbf{\textcolor{blue}{4.69}} & \cellcolor{gray!25}\textbf{\textcolor{blue}{68}} & \cellcolor{gray!25}\textbf{\textcolor{blue}{60}} & \cellcolor{gray!25}\textbf{\textcolor{blue}{52}}\\
			\bottomrule
		\end{tabular}
		\begin{tablenotes}    
			\footnotesize                              
			\item Note: Following established conventions, we report results with different precision formats across camera configurations: integers for panoramic settings and two decimal places for monocular settings. 
		\end{tablenotes}           
		\end{threeparttable} 	
	}\vspace{-4mm}
\end{table*}

\begin{table}[ht]
	\centering
	\caption{Experimental results on RxR-CE dataset.}
	\vspace{-3mm}
	\label{tab:rxr-ce}
	\resizebox{0.98\linewidth}{!}{
		%\begin{threeparttable}
			\begin{tabular}{@{}c|l|ccccc@{}}
				\toprule
				& \multicolumn{1}{c|}{} & \multicolumn{5}{c}{\textbf{Val Unseen}} \\ \cline{3-7}
				\multirow{-2}{*}{\textbf{Camera}} & \multicolumn{1}{c|}{\multirow{-2}{*}{\textbf{Methods}}} & \cellcolor{red!25}NE $\downarrow$  & \cellcolor{gray!25}SR & \cellcolor{gray!25}SPL & \cellcolor{gray!25}NDTW & \cellcolor{gray!25}SDTW \\ \midrule
				\multirow{8}{*}{\textbf{Monocular}}  
				&LAW~\cite{Raychaudhuri2021LanguageAlignedW}  & 10.87  & 8.0 & 8.0 & - & - \\
				&CM$^2$~\cite{Georgakis2022CrossmodalML}  & 8.98  & 14.4 & 9.2 & - & - \\
				&WS-MGMap~\cite{Chen2022WeaklySupervisedMM} & 9.83 & 15.0 & 12.1 & - & - \\
				& NaVid~\cite{zhang2024navid}  & \textbf{8.41} & 23.8 & 32.2 & - & - \\
				& A$^2$-Nav~\cite{chen20232} & - & 16.8 & 6.3 & - & - \\ \cmidrule(l){2-7} 
				& VLN-3DFF~\cite{wang2024sim} & 8.79 & 25.5 & 18.1 & - & - \\
				& \cellcolor{gray!10}VLN-3DFF*  & \cellcolor{gray!10}9.41 & \cellcolor{gray!10}26.66 & \cellcolor{gray!10}20.11 & \cellcolor{gray!10}42.91 & \cellcolor{gray!10}20.36 \\
				& \cellcolor{gray!25}\textbf{NavMorph} & \cellcolor{gray!25}\textcolor{blue}{8.85} & \cellcolor{gray!25}\textcolor{blue}{\textbf{30.76}} & \cellcolor{gray!25}\textcolor{blue}{\textbf{22.84}} & \cellcolor{gray!25}\textcolor{blue}{\textbf{44.19}} & \cellcolor{gray!25}\textcolor{blue}{\textbf{23.30}} \\  \midrule[0.7pt]
				\multirow{13}{*}{\textbf{Panoramic}} & LAW-Pano~\cite{Raychaudhuri2021LanguageAlignedW} & 11.04 & 10.0 & 9.0 & - & - \\
				& Seq2Seq~\cite{anderson2018vision} & 12.1 & 13.93 & 11.96 & 30.86 & 11.01 \\
				& CWP-CMA~\cite{Hong2022BridgingTG} & 8.76 & 26.59 & 22.16 & 47.05 & - \\
				& CWP-BERT~\cite{Hong2022BridgingTG} & 8.98 & 27.08 & 22.65 & 46.71 & - \\
				& AO-Planner~\cite{chen2024affordances} & 7.06 & 43.3 & 30.5 & 50.1 & - \\
				& Reborn~\cite{an20221st} & 5.98 & 48.60 & 42.05 & {63.35} & 41.82 \\\cmidrule(l){2-7} 
				%& HNR~\cite{wang2024lookahead} & 5.51 & 56.39 & 46.73 & 63.56 & 47.24 \\ 
				& ETPNav~\cite{an2024etpnav} & {5.64} & 54.79 & 44.89 & 61.90 & 45.33 \\ 
				& \cellcolor{gray!10}ETPNav* & \cellcolor{gray!10}5.96 & \cellcolor{gray!10}54.83 & \cellcolor{gray!10}44.62 & \cellcolor{gray!10}61.36 & \cellcolor{gray!10}44.87 \\
				&\cellcolor{gray!25}\textbf{NavMorph} & \cellcolor{gray!25}\textcolor{blue}{{5.80}}  & \cellcolor{gray!25}\textcolor{blue}{{56.23}}  & \cellcolor{gray!25}\textcolor{blue}{{46.39}}  & \cellcolor{gray!25}\textcolor{blue}{{63.23}}  & \cellcolor{gray!25}\textcolor{blue}{{46.98}}  \\\cmidrule(l){2-7}  
				& HNR~\cite{wang2024lookahead} & \textbf{5.51} & 56.39 & 46.73 & 63.56 & 47.24 \\ 
				& \cellcolor{gray!10}HNR* & \cellcolor{gray!10}5.75 & \cellcolor{gray!10}56.48 & \cellcolor{gray!10}46.62 & \cellcolor{gray!10}63.43 & \cellcolor{gray!10}47.38 \\
				&\cellcolor{gray!25}\textbf{NavMorph} & \cellcolor{gray!25}\textcolor{blue}{{5.70}}  & \cellcolor{gray!25}\textcolor{blue}{\textbf{58.02}}  & \cellcolor{gray!25}\textcolor{blue}{\textbf{48.98}}  & \cellcolor{gray!25}\textcolor{blue}{\textbf{64.77}}  & \cellcolor{gray!25}\textcolor{blue}{\textbf{48.85}}  \\ 
				%& ETPNav* & 6.03 & 53.38 & 42.37 & 59.55 & 43.57 \\
				%& \textbf{ETPNav-WM} & 5.82  & 54.87  & 43.98  & 60.41  & 44.58  \\ 
				\bottomrule
			\end{tabular}
			%	\begin{tablenotes}    
			%	\footnotesize
			%	\item Note: * indicates experimental results that we have reproduced in this work. 
			%\end{tablenotes}           
			%\end{threeparttable} 	
	}
	\vspace{-6mm}
\end{table}

\vspace{-2mm}
\section{Experiments}\vspace{-2mm}
We evaluate our model on R2R-CE and RxR-CE~\cite{krantz2020beyond} datasets, which transform discrete paths of R2R~\cite{anderson2018vision} and RxR~\cite{ku2020room} datasets into continuous environments.
Experiments and ablation studies show our effectiveness. %Please refer to the \textbf{Supplementary Material} for \textit{detailed experimental settings, more results and qualitative analysis}. %source code,

\vspace{-2mm}
\subsection{Experimental Setup}\vspace{-1mm}
\noindent\textbf{Datasets.} 
R2R-CE dataset~\cite{krantz2020beyond,anderson2018vision} comprises 5,611 shortest-path trajectories, divided across training, validation, and test splits. Each trajectory is paired with approximately three English instructions, with an average path length of 9.89 meters and a mean instruction length of 32 words. 
Based on similar scene splits, RxR-CE dataset~\cite{krantz2020beyond, ku2020room} is larger and more challenging, offering substantially more instructions in multiple languages, including English, Hindi, and Telugu, with an average of 120 words per instruction. 
Notably, agents in R2R-CE have a chassis radius of 0.10 meters and can slide along obstacles during navigation, while agents in RxR-CE, with a larger chassis radius of 0.18 meters, are restricted from sliding thus more prone to collisions.
%Notably, agents in the R2R-CE dataset have a chassis radius of 0.10 meters and can slide along obstacles during navigation, whereas agents in the RxR-CE dataset, with a larger chassis radius of 0.18 meters, are restricted from sliding, making them more susceptible to collisions.

\noindent\textbf{Evaluation Metrics.} 
We follow previous approaches~\cite{anderson2018vision, anderson2018evaluation, ilharco2019general} and adopt the standard metrics for evaluating VLN-CE agents, including TL (Trajectory Length), NE (Navigation Error), OSR (SR given Oracle stop policy), SR (Success Rate), SPL (Success weighted by Path Length), NDTW(Normalized Dynamic Time Warping), and SDTW (Success weighted by NDTW). 
%More details of these metrics can be found in \S~\ref{sec:metrics}.

\noindent\textbf{Implementation Details.}
For panoramic settings, we follow standard VLN-CE setups~\cite{Krantz2022Sim2SimTF,Hong2022BridgingTG}, where each location is represented by 12 RGB-D images captured at 30$^{\circ}$ intervals. We evaluate our approach with ETPNav\cite{an2024etpnav} and HNR\cite{wang2024lookahead} to verify our efficacy.
While monocular cameras offer advantages in real-world deployment, including lower cost, reduced size, and improved energy efficiency, we validate our method in a more practical setting. Specifically, we adopt VLN-3DFF~\cite{wang2024sim}, which adapts the original panoramic observation setting for deployment on real-world monocular robots, leveraging the pretrained 3D Feature Fields model~\cite{wang2024lookahead}.
To align with practical applications, we adhere to the online VLN setting established in~\cite{gao2024fast}, setting the batch size to 1 during evaluation.
All experiments were conducted using PyTorch framework~\cite{paszke2019pytorch} on a single NVIDIA RTX 3090 GPU. Please refer to \textbf{Supplementary Material} for detailed hyper-parameter settings.

\vspace{-1mm}
\subsection{Comparison with State-of-the-art VLN Models}\vspace{-1mm}
\noindent\textbf{R2R-CE.}
Table~\ref{tab:r2r-ce} presents a comparative analysis of our self-evolving world model against state-of-the-art methods on R2R-CE dataset. 
Compared to other monocular methods, our NavMorph consistently achieves strong performance across multiple metrics, particularly in unseen environments. 
Specifically, on the Val Unseen split, our model achieves remarkable gains of over 4$\%$ in both SR and SPL. The model's generalization capabilities are further validated on Test Unseen split, where NavMorph surpasses baseline VLN-3DFF by 1.6$\%$ in SPL and approximately 2$\%$ in SR.
%%In the panoramic setup, NavMorph consistently outperforms the baseline, underscoring its adaptability across visual setups. In addition, recent advanced methods have improved navigation performance by incorporating richer structural priors. For instance, BEVBert~\cite{an2023bevbert} leverages bird's-eye-view maps via multimodal pretraining, while HNR~\cite{wang2024lookahead} generates multi-level semantic features through pretrained neural radiance fields. As our work focuses on dynamic learning of the world model in VLN-CE, prioritizing its self-evolving capability over the use of more advanced feature representations, we do not directly compare with these methods. Recognizing potential benefits of enriched priors, we leave their integration for future work.
%As our work focuses on the self-evolutionary learning dynamics of world model in VLN-CE, we do not directly compare with these methods, though we recognize the potential benefits of enriched priors and leave their integration for future work.
 
Notably, our model achieves a marked reduction in trajectory length (TL) alongside increased SR under practical monocular settings, indicating more efficient navigation along optimized paths. This can be attributed to model’s robust ability to capture and predict latent environmental dynamics, enabling more informed decision-making through memory-enhanced representations.
With panoramic inputs, our model maintains comparable trajectory efficiency while further enhancing SR, likely due to the extensive visual context provided by panoramic views, which supplies sufficient information for base models to select well-informed paths.
These results underscore the efficacy of our NavMorph in improving navigation in continuous environments.

%While these approaches demonstrate the effectiveness of additional structural priors during pretraining, our work focuses primarily on investigating the self-evolutionary learning dynamics of world models in VLN tasks. Therefore, we consider these structural enhancements complementary to our approach and leave their integration for future exploration.
%Specifically, it achieves over 4$\%$ gains in SR and SPL on the Val Unseen split, two pivotal metrics for evaluating VLN performance. On the Test Unseen split, NavMorph further enhances generalization, surpassing the baseline VLN-3DFF by 1.6$\%$ on SPL and nearly 2$\%$ on SR.

%The richer visual information provided by panoramic views facilitates effective path selection, leading to smaller improvements in TL compared to the monocular setting.
%Compared to other monocular methods, NavMorph demonstrates consistent improvements across multiple metrics, particularly in unseen environments. Specifically, it achieves over 4$\%$ gains in both SR and SPL on the Val Unseen split, two pivotal metrics for evaluating VLN performance. On the Test Unseen split, NavMorph further enhances generalization, surpassing the baseline VLN-3DFF by 1.6$\%$ on SPL and nearly 2$\%$ on SR.

\noindent\textbf{RxR-CE.}
Table~\ref{tab:rxr-ce} compares NavMorph with existing methods on the RxR-CE dataset, illustrating its superior performance in both monocular and panoramic setups. In the monocular setting, NavMorph outperforms the previous state-of-the-art method VLN-3DFF by substantial margins of 4.1$\%$ in SR and 2.73$\%$ in SPL, highlighting its effectiveness in unseen environments. 
In panoramic setup, NavMorph achieves 58.02$\%$ SR and 48.98$\%$ SPL, performing competitively with specialized panoramic methods like ETPNav and HNR. The favorable performance across both NDTW (64.77) and SDTW (48.85) metrics further validates our model's capability to follow instructions faithfully while maintaining path fidelity.
These results underscore NavMorph’s robustness on the more challenging RxR-CE dataset, with its larger instruction set, multilingual support, and stricter navigation constraints. 

\vspace{-2mm}
\subsection{Further Remarks}\label{sec:4.3}\vspace{-2mm}
We perform ablation studies and other in-depth analysis of NavMorph on validation unseen set of R2R-CE dataset under the monocular setting, using VLN-3DFF as base model.

\noindent\textbf{Ablation Study of the proposed World Model.}
The ablation study presented in Table~\ref{tab:ablation} offers critical insights into the efficacy of our proposed world model. The removal of individual loss terms leads to moderate performance declines, underscoring their collective importance in capturing latent dynamics of the world model.
The visual embedding reconstruction loss $\ell_{re}$ and the action prediction loss $\ell_{ac}$, both optimized using NDTW, are crucial for navigation success. Removing $\ell_{re}$ leads to shorter path length (22.54 $\to$ 20.25) and reduced OSR (56.88$\%$ $\to$ 54.27$\%$), emphasizing the role of visual consistency in maintaining a temporally stable representation of surroundings. Similarly, omitting $\ell_{ac}$ degrades multiple metrics, as the constraint on predicted action sequence supports effective policy learning.
%, as the constraint on the predicted action sequence ensures alignment between intended and executed actions, thus supporting effective policy learning.
Moreover, the alignment between posterior and prior distributions is vital for maintaining consistency between learned latent dynamics and practical navigation scenarios, ensuring well-calibrated latent representations.
Additionally, to better assess contribution of self-evolution, we evaluate NavMorph \textit{w/o} SE, a variant where NavMorph operates with a fixed CEM module, which remains unchanged and does not apply refinements introduced in Eq.~\ref{eq:memory}. 
Rows 2-3 of Table~\ref{tab:ablation} highlight the impact of CEM in facilitating online adaptation through dynamic navigation insights aggregation.

Notably, the results validate not only the fundamental efficacy of our world model architecture—as evidenced by substantial performance gains even in the absence of self-evolution—but also complementary benefits of CEM-based adaptation, further enhancing navigation effectiveness. %This underscores that self-evolution serves as a valuable enhancement, strengthening the adaptability of a well-structured world model.
This underscores the collaborative nature of self-evolution and a well-structured world model, where self-evolution enhances adaptability while drawing support from the world model’s latent representations.
Extended experiments on self-evolution are provided in \textbf{Supplymentary Material}.

\begin{table}\centering
	\caption{Ablation Study of the proposed World Model.}
	\vspace{-3mm}
	\label{tab:ablation}
	\resizebox{0.99\linewidth}{!}{
		%\Huge
		\begin{threeparttable} 
			\begin{tabular}{@{}l|ccccccc@{}}
				\toprule
				\multicolumn{1}{c|}{\multirow{2}{*}{\textbf{Methods}}} &  \multicolumn{7}{c}{\textbf{R2R-CE Val Unseen}} \\ \cmidrule(l){2-8} 
				\multicolumn{1}{c|}{} & \cellcolor{red!25}TL $\downarrow$ & \cellcolor{red!25}NE $\downarrow$ & \cellcolor{gray!25}OSR & \cellcolor{gray!25}SR & \cellcolor{gray!25}SPL & \cellcolor{gray!25}NDTW & \cellcolor{gray!25}SDTW  \\ \midrule
				\multicolumn{1}{l|}{Base model} & 26.16 & 6.05 & 54.92 & 43.77 & 29.39 & 40.94 & 29.30 \\ \midrule
				%\multicolumn{1}{l|}{NavMorph-SE} & 23.33 & 5.77 & 56.12 & 46.87 & 32.56 & 44.42 & 32.16 \\
				\multicolumn{1}{l|}{\cellcolor{gray!25}{\textbf{NavMorph}}} &  \cellcolor{gray!25}{22.54} & \cellcolor{gray!25}{\textbf{5.75}} & \cellcolor{gray!25}{56.88} & \cellcolor{gray!25}{\textbf{47.91}} & \cellcolor{gray!25}{\textbf{33.22}} & \cellcolor{gray!25}{44.86} & \cellcolor{gray!25}{\textbf{32.73}}\\
				\multicolumn{1}{l|}{\quad\quad\quad \textit{w/o} $\ell_{re}$} & \textbf{20.25} & 5.85 & 54.27 & 45.51 & 32.38 & \textbf{45.92} & 32.21 \\
				\multicolumn{1}{l|}{\quad\quad\quad \textit{w/o} $\ell_{ac}$} & 25.14 & 5.96 & 56.06 & 44.81 & 31.22 & 43.04 & 30.49 \\
				\multicolumn{1}{l|}{\quad\quad\quad \textit{w/o} $\ell_{kl}$} & 25.69 & 6.30 & \textbf{57.15} & 44.10 & 30.44 & 41.20 & 29.91 \\ \midrule[0.7pt]
				\multicolumn{1}{l|}{NavMorph \textit{w/o} SE} & 23.34 & 5.92 & 55.19 & 45.08 & 31.19 & 42.70 & 30.85 \\\bottomrule
			\end{tabular}
			\begin{tablenotes}    
				\footnotesize               
				\item Note: `NavMorph \textit{w/o} SE' denotes NavMorph does not perform self-evolution. Last three rows denote NavMorph pretrained with $\mathcal{L}$ without specific loss.
			\end{tablenotes}           
		\end{threeparttable}  
	}\vspace{-2mm}
\end{table}

\begin{table}\centering
	\caption{Ablation Study on Contextual Evolution Memory (CEM).}
	\vspace{-3mm}
	\label{tab:ablation-de}
	\resizebox{0.98\linewidth}{!}{
		\begin{threeparttable} 
			\begin{tabular}{@{}c|c|ccccccc|c@{}}
				\toprule
				\multicolumn{2}{c|}{\multirow{2}{*}{\textbf{Methods}}} & \multicolumn{7}{c|}{\textbf{R2R-CE Val Unseen}} & \multirow{2}{*}{\textbf{\begin{tabular}[c]{@{}c@{}}Testing\\ Time*\end{tabular}}} \\ \cmidrule(lr){3-9}
				\multicolumn{2}{c|}{} & \cellcolor{red!25}TL $\downarrow$ & \cellcolor{red!25}NE $\downarrow$ & \cellcolor{gray!25}OSR & \cellcolor{gray!25}SR & \cellcolor{gray!25}SPL & \cellcolor{gray!25}NDTW & \cellcolor{gray!25}SDTW &  \\ \midrule
				\multicolumn{2}{c|}{Base Model} & 26.16 & 6.05 & 54.92 & 43.77 & 29.39 & 40.94 & 29.30 & 20.53s \\ \midrule
				\multirow{2}{*}{NavMorph} & \textit{LSTM-based} & 25.17 & 6.11 & 54.54 & 43.67 & 29.81 & 41.71 & 29.83 & 44.56s \\
				& \cellcolor{gray!25}\textbf{\textit{CEM-based}} & \cellcolor{gray!25}\textbf{{22.54}} & \cellcolor{gray!25}\textbf{{5.75}} & \cellcolor{gray!25}\textbf{{56.88}} & \cellcolor{gray!25}\textbf{{47.91}} & \cellcolor{gray!25}\textbf{{33.22}} & \cellcolor{gray!25}\textbf{{44.86}} & \cellcolor{gray!25}\textbf{{32.73}} & \cellcolor{gray!25}{21.22s} \\ \bottomrule
			\end{tabular}
			\begin{tablenotes}    
				\footnotesize               
				\item * Note: The last column displays the average execution time of the agnet for a single instruction, calculated on the Validation Unseen set of R2R-CE dataset. Best results are shown in bold. 
			\end{tablenotes}           
		\end{threeparttable} 
	}
	\vspace{-6mm}
\end{table}

\noindent\textbf{Ablation Study on CEM.}
In our model, CEM serves as the core module for self-evolution, enabling the agent to retain and refine historical scene information for improved adaptability in navigation tasks. Unlike conventional recurrent architectures (\textit{e.g.}, RNNs, LSTMs) that rely on gradient-based backpropagation to update model parameters for adaptation during online testing, CEM employs a forward iterative update mechanism, efficiently integrating new observations without incurring excessive computational costs.
To further validate this design choice, we conduct an ablation study under identical evaluation settings, comparing our \textit{CEM-based} self-evolution mechanism with \textit{LSTM-based} alternative. In the \textit{LSTM-based} mechanism, self-evolution is governed by an LSTM network, where optimization is performed via gradient backpropagation~\cite{gao2024fast} during online testing, and latent states are recursively updated through recurrent processing: $\tilde{\boldsymbol{h}}_t = LSTM(\boldsymbol{h}_t, \boldsymbol{o}_t).$

As shown in Table~\ref{tab:ablation-de}, our NavMorph with the CEM-based mechanism outperforms the LSTM-based approach across all key metrics, demonstrating superior adaptability in unseen environments. Notably, while both methods improve generalization through self-evolution, the LSTM-based approach incurs a $2.1\times$ increase in testing time due to its reliance on gradient-based optimization.
These results underscore the efficacy of our forward-update mechanism in balancing generalization and efficiency in VLN-CE tasks.
% \noindent\textbf{Different Top-$K$ Scene-related Visual Features retrieved from $\mathcal{M}$.}
%In our model, CEM serves as the core module for self-evolution, allowing the agent to retain and refine historical scene information for better adaptation in navigation tasks. Unlike conventional recurrent architectures (\textit{e.g.}, RNNs, LSTMs) that rely on gradient-based backpropagation to incorporate new environmental information into historical states, CEM employs a forward iterative update mechanism to efficiently integrate new observations without excessive computational cost.

\begin{figure}
	\centering
	\includegraphics[width=0.9\linewidth]{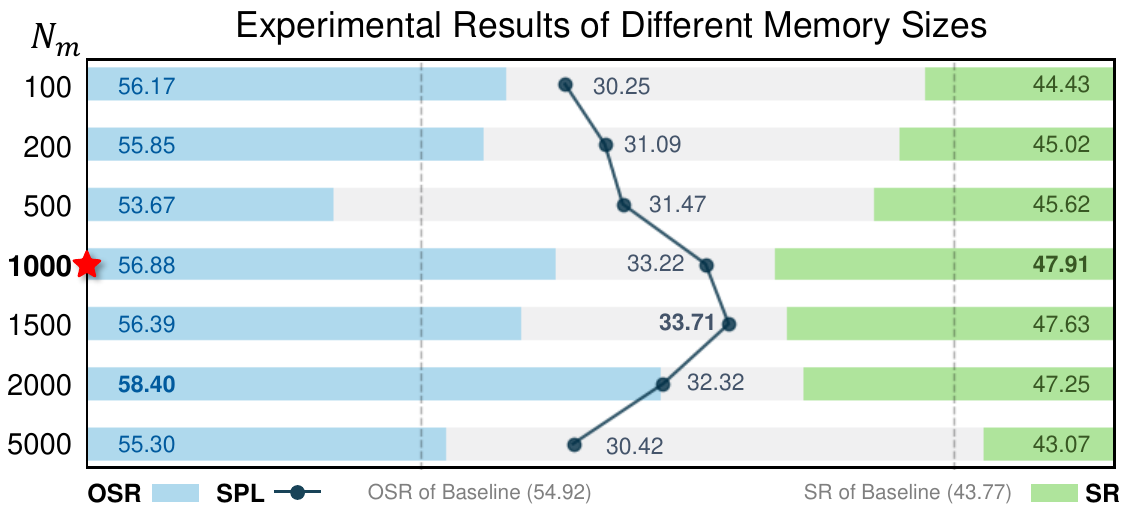}\vspace{-3mm}
	\caption{Experimental Results of Different Memory Sizes $N_m$.}
	\label{fig:size}
	\vspace{-3mm}
\end{figure}

\noindent\textbf{Size of Contextual Evolution Memory $N_m$.}
%The ablation study results indicate that the model’s performance across OSR, SR, and SPL metrics improves with increasing memory size, reaching optimal values at $N_m=1000$. This memory size strikes a balance between capturing essential scene-contextual information and maintaining efficient navigation performance. Beyond this point, further increases in memory size tend to slightly reduce effectiveness across these metrics, suggesting that an excessively large memory may introduce noise or reduce generalization capability in unseen environments. Overall, a moderate memory size of 1000 provides the most robust outcomes in terms of accuracy and path efficiency.
We conduct comprehensive ablation studies by varying the memory size $N_m$ within our Contextual Evolution Memory (CEM) from 100 to 5000. As shown in Figure~\ref{fig:size}, the model achieves optimal performance with $N_m = 1000$, where SR peaks at 47.91$\%$ (a 4.14$\%$ improvement over baseline), OSR reaches 56.88$\%$, and SPL attains 33.22$\%$. Interestingly, we observe that both smaller and larger memory sizes lead to performance degradation. This suggests that insufficient memory capacity constrains the model's ability to store essential contextual information, while excessive memory potentially introduces noise and redundancy that may interfere with effective decision-making. The results empirically validate a memroy size of $1000$ within CEM as the optimal balance between contextual retention and adaptability, enhancing the agent’s navigational robustness.

\begin{table} \centering
	\caption{Comparison with SOTA Online VLN Method.}
	\vspace{-3mm}
	\label{tab:online}
	\resizebox{0.98\linewidth}{!}{
		\begin{threeparttable} 
			\begin{tabular}{@{}c|ccccccc|c@{}}
				\toprule
				\multirow{2}{*}{\textbf{Methods}} & \multicolumn{7}{c|}{\textbf{R2R-CE Val Unseen}} & \multirow{2}{*}{\begin{tabular}[c]{@{}c@{}}Time*\\ (s)\end{tabular}} \\
				& \cellcolor{red!25}TL $\downarrow$ & \cellcolor{red!25}NE $\downarrow$ & \cellcolor{gray!25}OSR & \cellcolor{gray!25}SR & \cellcolor{gray!25}SPL & \cellcolor{gray!25}NDTW & \cellcolor{gray!25}SDTW &  \\ \midrule
				Base model & 26.16 & 6.05 & 54.92 & 43.77 & 29.39 & 40.94 & 29.30 & 20.53 \\ \midrule
				%\quad + FSTTA & 28.25 & 6.05 & 52.69 & 43.77 & 29.63 & 42.76 & 29.34 & 1342.7 \\\midrule
				FSTTA~\cite{gao2024fast} & 28.25 & 6.67 & 55.41 & 43.94 & 29.63 & 42.76 & 29.34 & 27.34 \\
				\rowcolor{gray!25}NavMorph & \textbf{22.54} & \textbf{5.75} & \textbf{56.88} & \textbf{47.91} & \textbf{33.22} & \textbf{44.86} & \textbf{32.73} & 21.22 \\ \bottomrule
			\end{tabular}
			\begin{tablenotes}    
				\footnotesize               
				\item * Note that the last column displays the average execution time of the VLN agent for a single instruction, calculated on the Validation Unseen set of R2R-CE.
			\end{tablenotes}           
		\end{threeparttable} 
	}
	\vspace{-6mm}
\end{table}

\noindent\textbf{Discussion.}
Astute readers may notice that our evolving world model accumulates scene-specific information from test environments stored as memory information during online testing.
While this memory mechanism operates without access to ground truth actions of testing samples, it may potentially raise concerns about evaluation fairness.
Recent research~\cite{gao2024fast} has established the practical significance of online test-time adaptation in VLN tasks, demonstrating its value for real-world deployments where agents must continuously adapt to novel environments. 
As shown in Table~\ref{tab:online}, our self-evolving world model, which employs memory-based forward updates, significantly outperforms gradient-based test-time adaptation method, FSTTA~\cite{gao2024fast}, in both navigation accuracy and efficiency. 
In fact, Rows 2-3 of Table~\ref{tab:ablation} also demonstrate the effectiveness of our approach in leveraging accumulated information for efficient adaptation while maintaining robust performance.

\vspace{-2mm}
\section{Conclusion}\vspace{-2mm}
This paper presents NavMorph, a novel self-evolving world model architecture that addresses the challenges of vision-and-language navigation in continuous environments. At its core, NavMorph integrates World-aware Navigator and Foresight Action Planner to model environmental dynamics in latent space, enabling effective adaptation to diverse online scenarios through cross-episodic learning by the introduced Contextual Evolution Memory mechanism. Extensive experiments demonstrate the effectiveness of our approach, suggesting valuable directions for advancing adaptive navigation tasks in complex dynamic environments. 

An open challenge is to design a navigational reward function directly from ground truth, enabling explicit planning within world models to better balance task objectives with environmental constraints, thus enhancing long-term decision-making. 
Another promising direction lies in developing a modular world model framework with standardized components, enabling flexible composition and evolution of agent capabilities through systematic modular abstractions.

\section*{Acknowledgment}
This work was supported in part by the National Key Research and Development Plan of China under Grant 2023YFC3310700, in part by the National Natural Science Foundation of China under Grants 62036012, U21B2044, 62236008, 62472422, and U2333215, and in part by Beijing Natural Science Foundation under Grant 4242051.

{
	\small
	\bibliographystyle{ieeenat_fullname}
	\bibliography{main}
}
\clearpage
\setcounter{page}{1}
\maketitlesupplementary

\S~\ref{sec:ELBO} provides a detail lower bound derivation for the defined loss $\mathcal{L}_{W}$. \S~\ref{sec:metrics} presents more details about evaluation metrics. Implementation details for experiments are provided in \S~\ref{sec:detail}, and further comparisons against state-of-the-art methods are shown in \S~\ref{sec:exper}. Finally, we present several visualizations for qualitative analysis in \S~\ref{sec:quali}.

\section{Lower Bound Derivation}\label{sec:ELBO}
Following the predictive network described in $\S$ 3 of our main paper, the joint probability distribution for our proposed world model can be factorized as:
\begin{align}
	\begin{split}
		p&(\boldsymbol{h}_{1:T}  , \boldsymbol{s}_{1:T},\boldsymbol{x}_{1:T+T_p}\boldsymbol{a}_{1:T+T_p}) = \\& \prod_{t=1}^{T} p(\boldsymbol{h}_{t},\boldsymbol{s}_{t}|\boldsymbol{h}_{t-1},\boldsymbol{s}_{t-1}, \boldsymbol{a}_{t-1}) p(\boldsymbol{x}_{t}, \boldsymbol{a}_{t}|\boldsymbol{h}_{t},\boldsymbol{s}_{t})
		\\& \prod_{j=1}^{T_p} p(\boldsymbol{h}_{T+j},\boldsymbol{s}_{T+j}|\boldsymbol{h}_{T},\boldsymbol{s}_{T}, \boldsymbol{a}_{T+j-1}) p(\boldsymbol{x}_{T+j}, \boldsymbol{a}_{T+j}|\boldsymbol{h}_{T},\boldsymbol{s}_{T}),
	\end{split}
\end{align}

For the first item for step 1 to $T$, we have:
\begin{align}
	\begin{split}
		p(\boldsymbol{h}_{t},\boldsymbol{s}_{t}|\boldsymbol{h}_{t-1},&\boldsymbol{s}_{t-1}, \boldsymbol{a}_{t-1}) = \\
		& p(\boldsymbol{h}_t|\boldsymbol{h}_{t-1},\boldsymbol{s}_{t-1}) p(\boldsymbol{s}_t|\boldsymbol{h}_{t},\boldsymbol{a}_{t-1}), 
	\end{split}
\end{align}
\begin{align}
	p(\boldsymbol{x}_{t}, \boldsymbol{a}_{t}|\boldsymbol{h}_{t},\boldsymbol{s}_{t}) = p(\boldsymbol{x}_t|\boldsymbol{h}_{t},\boldsymbol{s}_{t}) p(\boldsymbol{a}_t|\boldsymbol{h}_{t},\boldsymbol{s}_{t}),
\end{align}
%&p(\boldsymbol{h}_{j},\boldsymbol{s}_{j}|\boldsymbol{h}_{T},\boldsymbol{s}_{T}, \boldsymbol{a}_{j-1}) = p(\boldsymbol{h}_j|\boldsymbol{h}_{T},\boldsymbol{s}_{T}) p(\boldsymbol{s}_j|\boldsymbol{h}_{T},\boldsymbol{s}_{T}), \\
%&p(\boldsymbol{x}_{j}, \boldsymbol{a}_{j}|\boldsymbol{h}_{T},\boldsymbol{s}_{T}) = p(\boldsymbol{x}_j|\boldsymbol{h}_{T},\boldsymbol{s}_{T}) p(\boldsymbol{a}_j|\boldsymbol{h}_{T},\boldsymbol{s}_{T}),

Given that $\boldsymbol{h}_{t}$ is deterministic as discussed earlier, we have $p(\boldsymbol{h}_t|\boldsymbol{h}_{t-1},\boldsymbol{s}_{t-1})=\delta \left(\boldsymbol{h}_t-f_{\theta}(\boldsymbol{h}_{t-1},\boldsymbol{s}_{t-1}) \right)$.
Therefore, we need to infer the latent variables $\boldsymbol{s}_{1:T}$. Since no observations are available during the prediction phase $[T+1:T+T_p]$, the inference process focuses on maximizing the marginal likelihood over the observed data $p(\boldsymbol{x}_{1:T},\boldsymbol{a}_{1:T})$. Based on deep variational inference, we introduce a variational distribution $q_{H,S}$ and factorize as follows, for we assume that independence of $(\boldsymbol{x}_{1:T},\boldsymbol{a}_{1:T})$ given $(\boldsymbol{o}_{1:T},\boldsymbol{a}_{1:T-1})$:
\begin{align}\label{eq:qhs}
	\begin{split}
		q_{H,S} &\triangleq q(\boldsymbol{h}_{1:T+T_p},\boldsymbol{s}_{1:T+T_p}| \boldsymbol{o}_{1:T}, \boldsymbol{x}_{1:T+T_p}, \boldsymbol{a}_{1:T+T_p}) \\
		&\triangleq q(\boldsymbol{h}_{1:T},\boldsymbol{s}_{1:T}| \boldsymbol{o}_{1:T}, \boldsymbol{x}_{1:T}, \boldsymbol{a}_{1:T}) \\
		&\triangleq q(\boldsymbol{h}_{1:T},\boldsymbol{s}_{1:T}| \boldsymbol{o}_{1:T}, \boldsymbol{a}_{1:T-1}) \\
		& = \prod_{t=1}^{T} q(\boldsymbol{h}_{t} |\boldsymbol{h}_{t-1}, \boldsymbol{s}_{t-1}) q(\boldsymbol{s}_{t} | \boldsymbol{o}_{1:t}, \boldsymbol{a}_{1:t-1}),
	\end{split}
\end{align}
with $q(\boldsymbol{h}_{t} | \boldsymbol{h}_{t-1}, \boldsymbol{s}_{t-1})=p(h_{t} | \boldsymbol{h}_{t-1}, \boldsymbol{s}_{t-1})$ and $q_1(\boldsymbol{h}_1)=\delta(\boldsymbol{0})$. The Kullback-Leibler (KL) divergence between the prior and posterior distributions can be calculated as:
\begin{align}\label{eq:kl}
	\begin{split}
		& D_{\mathrm{KL}}( q(\boldsymbol{h}_{1:T}, \boldsymbol{s}_{1:T} | \boldsymbol{o}_{1:T}, \boldsymbol{x}_{1:T+T_p}, \boldsymbol{a}_{1:T+T_p}) \\
		& \qquad \qquad \parallel p(\boldsymbol{h}_{1:T}, \boldsymbol{s}_{1:T} | \boldsymbol{x}_{1:T+T_p}, \boldsymbol{a}_{1:T+T_p})) \\
		&=  \mathbb{E}_{\boldsymbol{h}_{1:T}, \boldsymbol{s}_{1:T} \sim q_{H,S}} \left[ \log \frac{q(\boldsymbol{h}_{1:T}, \boldsymbol{s}_{1:T} | \boldsymbol{o}_{1:T}, \boldsymbol{x}_{1:T+T_p}, \boldsymbol{a}_{1:T+T_p})}{p(\boldsymbol{h}_{1:T}, \boldsymbol{s}_{1:T} |\boldsymbol{x}_{1:T+T_p}, \boldsymbol{a}_{1:T+T_p})} \right]\\
		&=  \mathbb{E}_{\boldsymbol{h}_{1:T}, \boldsymbol{s}_{1:T} \sim q_{H,S}} \left[ \log \frac{q(\boldsymbol{h}_{1:T}, \boldsymbol{s}_{1:T} | \boldsymbol{o}_{1:T}, \boldsymbol{x}_{1:T+T_p}, \boldsymbol{a}_{1:T+T_p})}{p(\boldsymbol{h}_{1:T}, \boldsymbol{s}_{1:T} |\boldsymbol{x}_{1:T+T_p}, \boldsymbol{a}_{1:T+T_p})} \right]\\
		& =  \mathbb{E}_{\boldsymbol{h}_{1:T}, \boldsymbol{s}_{1:T} \sim q_{H,S}} \\
		&\left[ \log \frac{q(\boldsymbol{h}_{1:T}, \boldsymbol{s}_{1:T} | \boldsymbol{o}_{1:T}, \boldsymbol{x}_{1:T+T_p}, \boldsymbol{a}_{1:T+T_p})p( \boldsymbol{x}_{1:T+T_p}, \boldsymbol{a}_{1:T+T_p})}{p(\boldsymbol{h}_{1:T}, \boldsymbol{s}_{1:T} | \boldsymbol{x}_{1:T+T_p}, \boldsymbol{a}_{1:T+T_p}) p( \boldsymbol{h}_{1:T}, \boldsymbol{s}_{1:T})} \right]\\
		&=  \log p(\boldsymbol{x}_{1:T+T_p}, \boldsymbol{a}_{1:T+T_p}) \\
		& - \mathbb{E}_{\boldsymbol{h}_{1:T}, \boldsymbol{s}_{1:T} \sim q_{H,S}} \left[ \log p(\boldsymbol{x}_{1:T+T_p}, \boldsymbol{a}_{1:T+T_p} | \boldsymbol{h}_{1:T}, \boldsymbol{s}_{1:T}) \right]\\
		& + D_{\mathrm{KL}} \left( q(\boldsymbol{h}_{1:T}, \boldsymbol{s}_{1:T} | \boldsymbol{o}_{1:T}, \boldsymbol{x}_{1:T+T_p}, \boldsymbol{a}_{1:T+T_p}) \, \| \, p(\boldsymbol{h}_{1:T}, \boldsymbol{s}_{1:T}) \right).
	\end{split}
\end{align}

Since $D_{KL} \geq 0$, the left side of Eq.~\eqref{eq:kl} should be non-negative. Based on Jensen's inequality~\cite{jensen1906fonctions}, a variational lower bound on the log evidence can be obtained as follows:
\begin{align}\label{eq:lowerbound}
	\begin{split}
		\log p& (  \boldsymbol{x}_{1:T+T_p},\boldsymbol{a}_{1:T+T_p}) \geq \\ 
		& \mathbb{E}_{\boldsymbol{h}_{1:T}, \boldsymbol{s}_{1:T} \sim q_{H,S}} \left[ \log p(\boldsymbol{x}_{1:T+T_p}, \boldsymbol{a}_{1:T+T_p} | \boldsymbol{h}_{1:T}, \boldsymbol{s}_{1:T}) \right]-  \\
		&  D_{\mathrm{KL}} \left( q(\boldsymbol{h}_{1:T}, \boldsymbol{s}_{1:T} | \boldsymbol{o}_{1:T}, \boldsymbol{x}_{1:T+T_p}, \boldsymbol{a}_{1:T+T_p}) \, \| \, p(\boldsymbol{h}_{1:T}, \boldsymbol{s}_{1:T}) \right). 
	\end{split}
\end{align}

As for the first term of the lower bound in Eq.~\eqref{eq:lowerbound}:
\begin{align}\label{eq:likelihood}
	\begin{split}
		&\mathbb{E}_{\boldsymbol{h}_{1:T}, \boldsymbol{s}_{1:T} \sim q_{H,S}} \left[ \log p(\boldsymbol{x}_{1:T+T_p}, \boldsymbol{a}_{1:T+T_p} | \boldsymbol{h}_{1:T}, \boldsymbol{s}_{1:T}) \right]\\
		&= \mathbb{E}_{\boldsymbol{h}_{1:T}, \boldsymbol{s}_{1:T} \sim q_{H,S}} \left[ \log \prod_{t=1}^{T} p(\boldsymbol{x}_{t} | \boldsymbol{h}_{t}, \boldsymbol{s}_{t}) p(\boldsymbol{a}_{t} | \boldsymbol{h}_{t}, \boldsymbol{s}_{t}) \right.\\
		&\quad \left. \prod_{j=1}^{T_p} p(\boldsymbol{h}_{T+j},\boldsymbol{s}_{T+j}|\boldsymbol{h}_{T},\boldsymbol{s}_{T}, \boldsymbol{a}_{T+j-1}) p(\boldsymbol{x}_{T+j}, \boldsymbol{a}_{T+j}|\boldsymbol{h}_{T},\boldsymbol{s}_{T}) \right] \\
		&= \sum_{t=1}^{T} \mathbb{E}_{\boldsymbol{h}_{1:t}, \boldsymbol{s}_{1:t} \sim q(\boldsymbol{h}_{t}, \boldsymbol{s}_{t} | \boldsymbol{o}_{1:t}, \boldsymbol{a}_{1:t-1})}[ \log p(\boldsymbol{x}_{t} | \boldsymbol{h}_{t}, \boldsymbol{s}_{t}) \\
		& \quad +  \log p(\boldsymbol{a}_{t} | \boldsymbol{h}_{t}, \boldsymbol{s}_{t})]  + \sum_{j=1}^{T_p} \mathbb{E}_{\boldsymbol{h}_{T}, \boldsymbol{s}_{T} \sim q(\boldsymbol{h}_{T}, \boldsymbol{s}_{T} | \boldsymbol{o}_{1:t}, \boldsymbol{a}_{1:t-1})} \\
		& \qquad [ \log p(\boldsymbol{x}_{T+j} | \boldsymbol{h}_{T}, \boldsymbol{s}_{T}) + \log p(\boldsymbol{a}_{T+j} | \boldsymbol{h}_{T}, \boldsymbol{s}_{T}) ],
	\end{split}
\end{align}
where Eq.~\eqref{eq:likelihood} is obtained by integrating over remaining latent variables $(\boldsymbol{h}_{t:1+T},\boldsymbol{s}_{t:1+T})$.

Regarding second term of the lower bound in Eq.~\eqref{eq:lowerbound}, since there are no observations available during the prediction phase $[T+1:T+T_p]$, the posterior distribution $q$ is no longer updated with new input information.  Consequently, it converges to the prior distribution, making the KL divergence between the posterior $q(\boldsymbol{h}_{T:T+T_p}, \boldsymbol{s}_{T:T+T_p})$ and the prior $p(\boldsymbol{h}_{T:T+T_p}, \boldsymbol{s}_{T:T+T_p})$ equal to zero. As a result, only the KL divergence for the observed phase $[1:T]$ needs to be considered, which can be calculated according to Eq.~\eqref{eq:qhs}: 
\begin{align}\label{eq:logfinal}
	\begin{split}  
		D&_{\mathrm{KL}} \left( q(\boldsymbol{h}_{1:T}, \boldsymbol{s}_{1:T} | \boldsymbol{o}_{1:T}, \boldsymbol{x}_{1:T+T_p}, \boldsymbol{a}_{1:T+T_p}) \, \| \, p(\boldsymbol{h}_{1:T}, \boldsymbol{s}_{1:T}) \right)\\
		\triangleq& D_{\mathrm{KL}} \left( q(\boldsymbol{h}_{1:T}, \boldsymbol{s}_{1:T} | \boldsymbol{o}_{1:T}, \boldsymbol{a}_{1:T-1}) \, \| \, p(\boldsymbol{h}_{1:T}, \boldsymbol{s}_{1:T}) \right) \\
		= & \mathbb{E}_{\boldsymbol{h}_{1:T}, \boldsymbol{s}_{1:T} \sim q_{H,S}} \left[ \log \frac{q(\boldsymbol{h}_{1:T}, \boldsymbol{s}_{1:T} | \boldsymbol{o}_{1:T}, \boldsymbol{a}_{1:T-1})}{p(\boldsymbol{h}_{1:T}, \boldsymbol{s}_{1:T})} \right] \\
		= & \int_{\boldsymbol{h}_{1:T}, \boldsymbol{s}_{1:T}} q(\boldsymbol{h}_{1:T}, \boldsymbol{s}_{1:T} | \boldsymbol{o}_{1:T}, \boldsymbol{a}_{1:T-1}) \\
		& \left( \log \frac{q(\boldsymbol{h}_{1:T}, \boldsymbol{s}_{1:T} | \boldsymbol{o}_{1:T}, \boldsymbol{a}_{1:T-1})}{p(\boldsymbol{h}_{1:T}, \boldsymbol{s}_{1:T})} \right) \, \mathrm{d}\boldsymbol{h}_{1:T} \, \mathrm{d}\boldsymbol{s}_{1:T} \\
		= & \int_{\boldsymbol{h}_{1:T}, \boldsymbol{s}_{1:T}} q(\boldsymbol{h}_{1:T}, \boldsymbol{s}_{1:T} | \boldsymbol{o}_{1:T}, \boldsymbol{a}_{1:T-1}) \\
		&\log \left[ \prod_{t=1}^{T} \frac{q(\boldsymbol{h}_{t} | \boldsymbol{h}_{t-1}, \boldsymbol{s}_{t-1}) q(\boldsymbol{s}_{t} | \boldsymbol{o}_{1:t}, \boldsymbol{a}_{1:t-1})}{p(\boldsymbol{h}_{t} | \boldsymbol{h}_{t-1}, \boldsymbol{s}_{t-1}) p(\boldsymbol{s}_{t} | \boldsymbol{h}_{t-1}, \boldsymbol{s}_{t-1})} \right] \mathrm{d}\boldsymbol{h}_{1:T} \, \mathrm{d}\boldsymbol{s}_{1:T} \\
		= &\int_{\boldsymbol{h}_{1:T}, \boldsymbol{s}_{1:T}} \left( \prod_{t=1}^{T} q(\boldsymbol{h}_{t} | \boldsymbol{h}_{t-1}, \boldsymbol{s}_{t-1}) q(\boldsymbol{s}_{t} | \boldsymbol{o}_{1:t}, \boldsymbol{a}_{1:t-1}) \right) \\
		& \left( \sum_{t=1}^{T} \log \frac{q(\boldsymbol{s}_{t} | \boldsymbol{o}_{1:t}, \boldsymbol{a}_{1:t-1})}{p(\boldsymbol{s}_{t} | \boldsymbol{h}_{t-1}, \boldsymbol{s}_{t-1})} \right) \mathrm{d}\boldsymbol{h}_{1:T} \, \mathrm{d}\boldsymbol{s}_{1:T}. \\
	\end{split}
\end{align}
Based on the above deduction, we iteratively integrate out each latent variable and, by recursively applying this process to the sum of logarithmic terms indexed by $t$, decompose the KL divergence into a summation over time steps.
\begin{align*}
	\begin{split}
		D&_{\mathrm{KL}} \left( q(\boldsymbol{h}_{1:T}, \boldsymbol{s}_{1:T} | \boldsymbol{o}_{1:T}, \boldsymbol{x}_{1:T}, \boldsymbol{a}_{1:T}) \, \| \, p(\boldsymbol{h}_{1:T}, \boldsymbol{s}_{1:T}) \right) \\
		= &\int_{\boldsymbol{h}_{1:T}, \boldsymbol{s}_{1:T}} \left( \prod_{t=1}^{T} q(\boldsymbol{h}_{t} | \boldsymbol{h}_{t-1}, \boldsymbol{s}_{t-1}) q(\boldsymbol{s}_{t} | \boldsymbol{o}_{1:t}, \boldsymbol{a}_{1:t-1}) \right) \\
		& \left(  \log \frac{q(\boldsymbol{s}_{1} | \boldsymbol{o}_{1})}{p(\boldsymbol{s}_{1})} + \sum_{t=2}^{T} \log \frac{q(\boldsymbol{s}_{t} | \boldsymbol{o}_{1:t}, \boldsymbol{a}_{1:t-1})}{p(\boldsymbol{s}_{t} | \boldsymbol{h}_{t-1}, \boldsymbol{s}_{t-1})} \right) \mathrm{d}\boldsymbol{h}_{1:T} \, \mathrm{d}\boldsymbol{s}_{1:T}\\
		= & \mathbb{E}_{\boldsymbol{s}_{1} \sim q(\boldsymbol{s}_{1} | \boldsymbol{o}_{1})} \left[ \log \frac{q(\boldsymbol{s}_{1} | \boldsymbol{o}_{1})}{p(\boldsymbol{s}_{1})} \right] \\
		& + \int_{\boldsymbol{h}_{1:T}, \boldsymbol{s}_{1:T}} \left( \prod_{t=1}^{T} q(\boldsymbol{h}_{t} | \boldsymbol{h}_{t-1}, \boldsymbol{s}_{t-1}) q(\boldsymbol{s}_{t} | \boldsymbol{o}_{1:t}, \boldsymbol{a}_{1:t-1}) \right) \\
		& \left( \sum_{t=2}^{T} \log \frac{q(\boldsymbol{s}_{t} | \boldsymbol{o}_{1:t}, \boldsymbol{a}_{1:t-1})}{p(\boldsymbol{s}_{t} | \boldsymbol{h}_{t-1}, \boldsymbol{s}_{t-1})} \right) \mathrm{d}\boldsymbol{h}_{1:T} \, \mathrm{d}\boldsymbol{s}_{1:T} 
	\end{split}
\end{align*}
\begin{align}\label{eq:klfinal}
	\begin{split}
		= & D_{\mathrm{KL}} \left( q(\boldsymbol{s}_{1} | \boldsymbol{o}_{1}) \, \| \, p(\boldsymbol{s}_{1}) \right) \\
		& + \int_{\boldsymbol{h}_{1:T}, \boldsymbol{s}_{1:T}} \left( \prod_{t=1}^{T} q(\boldsymbol{h}_{t} | \boldsymbol{h}_{t-1}, \boldsymbol{s}_{t-1}) q(\boldsymbol{s}_{t} | \boldsymbol{o}_{1:t}, \boldsymbol{a}_{1:t-1}) \right) \\
		& \left( \log \frac{q(\boldsymbol{s}_{2} | \boldsymbol{o}_{1:2}, \boldsymbol{a}_{1})}{p(\boldsymbol{s}_{2} | \boldsymbol{h}_{1}, \boldsymbol{s}_{1})} + \sum_{t=3}^{T} \log \frac{q(\boldsymbol{s}_{t} | \boldsymbol{o}_{1:t}, \boldsymbol{a}_{1:t-1})}{p(\boldsymbol{s}_{t} | \boldsymbol{h}_{t-1}, \boldsymbol{s}_{t-1})} \right) \mathrm{d}\boldsymbol{h}_{1:T} \, \mathrm{d}\boldsymbol{s}_{1:T} \\
		= & D_{\mathrm{KL}} \left( q(\boldsymbol{s}_{1} | \boldsymbol{o}_{1}) \, \| \, p(\boldsymbol{s}_{1}) \right) + \\
		& \mathbb{E}_{\boldsymbol{h}_{1}, \boldsymbol{s}_{1} \sim q(\boldsymbol{h}_{1}, \boldsymbol{s}_{1} | \boldsymbol{o}_{1})} \left[ D_{\mathrm{KL}} \left( q(\boldsymbol{s}_{2} | \boldsymbol{o}_{1:2}, \boldsymbol{a}_{1}) \, \| \, p(\boldsymbol{s}_{2} | \boldsymbol{h}_{1}, \boldsymbol{s}_{1}) \right) \right] \\
		& + \int_{\boldsymbol{h}_{1:T}, \boldsymbol{s}_{1:T}} \left( \prod_{t=1}^{T} q(\boldsymbol{h}_{t} | \boldsymbol{h}_{t-1}, \boldsymbol{s}_{t-1}) q(\boldsymbol{s}_{t} | \boldsymbol{o}_{1:t}, \boldsymbol{a}_{1:t-1}) \right) \\
		& \left( \sum_{t=3}^{T} \log \frac{q(\boldsymbol{s}_{t} | \boldsymbol{o}_{1:t}, \boldsymbol{a}_{1:t-1})}{p(\boldsymbol{s}_{t} | \boldsymbol{h}_{t-1}, \boldsymbol{s}_{t-1})} \right) \mathrm{d}\boldsymbol{h}_{1:T} \, \mathrm{d}\boldsymbol{s}_{1:T}\\
		= & \sum_{t=1}^{T} \mathbb{E}_{\boldsymbol{h}_{t-1}, \boldsymbol{s}_{t-1} \sim q(\boldsymbol{h}_{t-1}, \boldsymbol{s}_{t-1} | \boldsymbol{o}_{1:t-1}, \boldsymbol{a}_{1:t-2})}\\
		& \quad \left[ D_{\mathrm{KL}} \left( q(\boldsymbol{s}_{t} | \boldsymbol{o}_{1:t}, \boldsymbol{a}_{1:t-1}) \, \| \, p(\boldsymbol{s}_{t} | \boldsymbol{h}_{t-1}, \boldsymbol{s}_{t-1}) \right) \right]
	\end{split}
\end{align}
Combining Eq.~\eqref{eq:logfinal}, Eq.~\eqref{eq:klfinal} and Eq.~\eqref{eq:lowerbound}, the final lower bound can be obtained as follows:
\begin{align}
	\begin{split}
		\log p& (  \boldsymbol{x}_{1:T+T_p},\boldsymbol{a}_{1:T+T_p}) \geq \\ 
		& \sum_{t=1}^{T} \mathbb{E}_{\boldsymbol{h}_{1:t}, \boldsymbol{s}_{1:t} \sim q(\boldsymbol{h}_{t}, \boldsymbol{s}_{t} | \boldsymbol{o}_{1:t}, \boldsymbol{a}_{1:t-1})}[ \log p(\boldsymbol{x}_{t} | \boldsymbol{h}_{t}, \boldsymbol{s}_{t}) \\
		& +  \log p(\boldsymbol{a}_{t} | \boldsymbol{h}_{t}, \boldsymbol{s}_{t})]  + \sum_{j=1}^{T_p} \mathbb{E}_{\boldsymbol{h}_{T}, \boldsymbol{s}_{T} \sim q(\boldsymbol{h}_{T}, \boldsymbol{s}_{T} | \boldsymbol{o}_{1:t}, \boldsymbol{a}_{1:t-1})} \\
		& \quad [ \log p(\boldsymbol{x}_{T+j} | \boldsymbol{h}_{T}, \boldsymbol{s}_{T}) + \log p(\boldsymbol{a}_{T+j} | \boldsymbol{h}_{T}, \boldsymbol{s}_{T}) ],\\
		& - \sum_{t=1}^{T} \mathbb{E}_{\boldsymbol{h}_{t-1}, \boldsymbol{s}_{t-1} \sim q(\boldsymbol{h}_{t-1}, \boldsymbol{s}_{t-1} | \boldsymbol{o}_{1:t-1}, \boldsymbol{a}_{1:t-2})} \\
		& \qquad \left[ D_{\mathrm{KL}} \left( q(\boldsymbol{s}_{t} | \boldsymbol{o}_{1:t}, \boldsymbol{a}_{1:t-1}) \, \| \, p(\boldsymbol{s}_{t} | \boldsymbol{h}_{t-1}, \boldsymbol{s}_{t-1}) \right) \right]
	\end{split}
\end{align}

%%%%%%%%%%%%%%%%%%%%%%%%%%%%%%%%%%%%%%%%%%%%%%%%%%%%
%\section{Details of Online Adaptive Optimization}
%\label{sec:TTA}
%During the testing phase, we perform

%To split the supplementary pages from the main paper, you can use \href{https://support.apple.com/en-ca/guide/preview/prvw11793/mac#:~:text=Delete%20a%20page%20from%20a,or%20choose%20Edit%20%3E%20Delete).}{Preview (on macOS)}, \href{https://www.adobe.com/acrobat/how-to/delete-pages-from-pdf.html#:~:text=Choose%20%E2%80%9CTools%E2%80%9D%20%3E%20%E2%80%9COrganize,or%20pages%20from%20the%20file.}{Adobe Acrobat} (on all OSs), as well as \href{https://superuser.com/questions/517986/is-it-possible-to-delete-some-pages-of-a-pdf-document}{command line tools}.

%\section{Source Code}\label{sec:code}
%The code to reproduce our experimental results can be found in \href{https://github.com/Feliciaxyao/NavMorph}{NavMorph}. For more details, please refer to the README file inside our code.

%%%%%%%%%%%%%%%%%%%%%%%%%%%%%%%%%%%%%%%%%%%%%%%%%%%%

%%%%%%%%%%%%%%%%%%%%%%%%%%%%%%%%%%%%%%%%%%%%%%%%%%%%
\section{Evaluation Metrics for VLN-CE agents}
\label{sec:metrics}
We follow previous approaches~\cite{anderson2018vision, anderson2018evaluation,ilharco2019general} and adopt the standard metrics for evaluating VLN-CE agents:
\begin{itemize}[left=1em]
	\item TL (Trajectory length) measures the average length of the predicted navigation trajectories.
	\item NE (Navigation Error) measures the average distance (in meter) between the agent's final position in the predicted trajectory and the target in the ground truth.
	\item SR (Success Rate) is the proportion of the agent stopping
	in the predicted route within a threshold distance (set as 3 meters) of the goal
	in the reference route.
	\item OSR (Oracle Success Rate) is the proportion of the closest point in the predicted trajectory to the target in the reference trajectory within a threshold distance. 
	\item SPL (Success weighted by Path Length) )is a comprehensive metric method integrating SR and TL that takes both effectiveness and efficiency into account.
	\item NDTW (Normalized Dynamic Time Warping) measures the normalized cumulative distance between reference path and agent position.
	\item SDTW (Success weighted by normalized Dynamic Time
	Warping) is a comprehensive metric method integrating NDTW and SR that takes both path efficiency and task completion into account.
\end{itemize}

\section{Implementation Details}\label{sec:detail}
\textbf{The Baseline Framework.}
In conventional panoramic VLN-CE frameworks~\cite{an2024etpnav, wang2023gridmm}, the agent perceives its surroundings through multi-view RGB-D panoramas captured at 30-degree intervals at each timestep $t$. These panoramic observations are processed by a trained waypoint prediction module\cite{Hong2022BridgingTG} to identify navigable waypoints. The VLN model then encodes both the visual features of these waypoints and their spatial information (relative direction and distance) to construct a topological map. This map is subsequently integrated with the navigation instruction via the Cross-Modal Graph Transformer\cite{chen2022think, an2024etpnav}, which selects the optimal waypoint as the agent’s next navigation goal.

Monocular VLN-CE settings rely on a single RGB-D camera, which presents challenges in waypoint estimation due to the lack of full panoramic coverage. To address this, an enhanced waypoint predictor~\cite{wang2024sim} utilizes a semantic traversability map and 3D feature fields to infer viable waypoints, ensuring effective decision-making even with limited field-of-view.
%In conventional VLN-CE frameworks~\cite{an2024etpnav, wang2023gridmm}, the agent acquires multi-view observations comprising RGB-D panoramas captured at 30-degree intervals at each timestep $t$. These environmental inputs are processed by a trained waypoint prediction module~\cite{Hong2022BridgingTG} to identify viable navigable waypoints around the agent. The VLN model encodes the visual features of these waypoints alongside their spatial information (relative direction and distance) to construct a topological map. This map is then integrated with the navigation instruction using the Cross-Modal Graph Transformer~\cite{chen2022think,an2024etpnav}, which selects the optimal waypoint as the agent's next navigation goal. For monocular setups, an enhanced waypoint predictor~\cite{wang2024sim} leverages a semantic traversable map and 3D feature fields to predict waypoints.

\noindent\textbf{Model Configuration.}
Following the previous baseline model~\cite{an2024etpnav, wang2024sim}, we utilize CLIP-pretrained ViT-B/32~\cite{dosovitskiy2020image} for RGB feature extraction, while depth information is processed through a point-goal navigation pretrained ResNet-50~\cite{he2016deep}. The framework maintains encoder depths of 2, 9, and 4 layers for panoramic, textual, and cross-modal graph components respectively, aligned with~\cite{Hong2022BridgingTG, Georgakis2022CrossmodalML}.  Other hyperparameters are the same as LXMERT~\cite{tan2019lxmert} on the R2R-CE dataset and pre-trained RoBerta~\cite{liu2019roberta} for the multilingual RxR-CE dataset. The camera's HFOV is set to $90^{\circ}$ for R2R-CE and $79^{\circ}$ for RxR-CE.

\noindent\textbf{Experimental Details.}
NavMorph was trained over 10K episodes on the R2R-CE dataset and 20K episodes on the RxR-CE dataset, following the same initialization and training strategies as the pretrained baseline~\cite{wang2024sim,an2024etpnav}. 
The learning rate is set to $1 \times 10^{-5}$, while the weighting coefficient for loss function $\mathcal{L}$ is $\gamma = 10^{-3}$. Note that the weighting coefficients are heuristically adjusted to balance each loss term, ensuring they remain at the same order of magnitude based on initial values.
%for loss function are set heuristically, balancing each loss to same order of magnitude based on initial values.(\underline{\textit{Eq. (7) in main paper}})

At each timestep of a navigation task, the model predicts actions for $T_p = 2$ consecutive future states, starting from $t = 1$ (\textit{i.e.}, the next position after the agent's initial point). Accordingly, the observation window $T$ dynamically expands throughout the navigation process, increasing until the agent selects `stop' action or reaches the maximum step limit.
For input image dimensions $N_o \times h \times w \times c$, we set $1 \times 224 \times 224 \times 3$ for monocular settings and $12 \times 224 \times 224 \times 3$ for panoramic settings. The encoded visual embedding has a dimension of $d_x = 512$, while both scene-contextual features ($d_v$) and action embedding ($d_a$) are set to 768.
Our Contextual Evolution Memory (CEM) is initially randomized and progressively updated with informative scene-contextual features. These features are derived from panoramic visual representations extracted by the panoramic encoder during training, encapsulating comprehensive environmental information to enhance navigation. The memory size $N_m$ is set to 1000.
%Our Contextual Evolution Memory (CEM) is initialized with informative scene-contextual features derived from panoramic visual features extracted by the panoramic encoder during training. These features encapsulate comprehensive environmental information, serving as contextual representations to enhance navigation. The memory size $N_m$ is set to 1000.
For the monocular setting, $K$ is set to 16 for top-$K$ retrieval with update factors $\alpha = \beta = 0.7$.
For panoramic setting, $K$ is set to 10 for top-$K$ retrieval with update factors $\alpha = \beta = 0.9$.
%The dimensions of input images $N_o \times h \times w \times c$ are set to $1 \times 224 \times 224 \times 3$ for monocular settings and $12 \times 224 \times 224 \times 3$ for panoramic settings, while the encoded visual embeddings are set to $d_x = 512$. Both the scene-contextual feature dimension $d_v$ and action embedding dimension $d_a$ are 768. 
%And our CEM is initialized with informative scene-contextual features, which are derived from panoramic visual features extracted by the panoramic encoder during the training process. These features encapsulate comprehensive environmental information and serve as contextual representations to support effective navigation. The memory size $N_m$ is set to 1000. 

%We adopt full-training paradigm rather than short-training, with initialization and training strategies aligned with VLN-3DFF, given a consistent backbone.Specifically, 

\noindent\textbf{Working Modes.}
During the \textbf{\textit{training}} phase, NavMorph operates through two core components: the world-aware navigator, which executes navigation actions for VLN-CE tasks, and the foresight action planner, which performs imaginative rollouts for future $T_p$ steps. This collaborative framework enables the model to learn effective navigation strategies while simultaneously refining its latent state representation capabilities.
During the \textbf{\textit{online testing}} phase, the world-aware navigator performs navigation planning by leveraging the predicted future actions generated by the foresight action planner as guidance. Specifically, the navigator evaluates each candidate waypoint by assigning navigation scores based on the learned policy, which are subsequently refined according to their proximity to the predicted trajectory points. This weighting strategy prioritizes candidates closer to the predicted path, seamlessly integrating the foresight planner's predictions into final navigation decision-making process.

\begin{table*}[thpb] \centering
	\caption{Experimental results on R2R-CE dataset. Results better than base model are shown in \textcolor{blue}{blue}. Best results for the panoramic and monocular settings are each highlighted in \textbf{bold}. * indicates experimental results that we have reproduced in this work.}
	\vspace{-3mm}
	\label{tab:r2r-ce-full}
	\resizebox{0.99\textwidth}{!}{
		%\Huge
		\begin{threeparttable}
			\begin{tabular}{@{}c|l|ccccccccccccccc@{}}
				\toprule
				& \multicolumn{1}{c|}{} & \multicolumn{5}{c|}{\textbf{Val Seen}} & \multicolumn{5}{c|}{\textbf{Val Unseen}} & \multicolumn{5}{c}{\textbf{Test Unseen}} \\  
				\multirow{-2}{*}{\textbf{Camera}} & \multicolumn{1}{c|}{\multirow{-2}{*}{\textbf{Methods}}} & \cellcolor{red!25}TL $\downarrow$ & \cellcolor{red!25}NE $\downarrow$ & \cellcolor{gray!25}OSR & \cellcolor{gray!25}SR & \cellcolor{gray!25}SPL & \cellcolor{red!25}TL $\downarrow$ & \cellcolor{red!25}NE $\downarrow$ & \cellcolor{gray!25}OSR & \cellcolor{gray!25}SR & \cellcolor{gray!25}SPL & \cellcolor{red!25}TL $\downarrow$ & \cellcolor{red!25}NE $\downarrow$ & \cellcolor{gray!25}OSR & \cellcolor{gray!25}SR & \cellcolor{gray!25}SPL \\ \midrule
				& LAW~\cite{Raychaudhuri2021LanguageAlignedW}& \textbf{9.34} & 6.35 & 49 & 40 & \multicolumn{1}{c|}{37} & \textbf{8.89} & 6.83 & 44 & 35 & \multicolumn{1}{c|}{31} & \textbf{9.67} & 7.69 & 28 & 38 & 25 \\
				\multicolumn{1}{c|}{} & CM$^2$~\cite{Georgakis2022CrossmodalML}  & 12.05 & 6.10 & 50.7 & 42.9 & \multicolumn{1}{c|}{34.8} & 11.54 & 7.02 & 41.5 & 34.3 & \multicolumn{1}{c|}{27.6} & 13.90 & 7.70 & 39 & 31 & 24 \\
				\multicolumn{1}{c|}{} & WS-MGMap~\cite{Chen2022WeaklySupervisedMM}& 10.12 & 5.65 & 51.7 & 46.9 & \multicolumn{1}{c|}{\textbf{43.4}} & 10.00 & 6.28 & 47.6 & 38.9 & \multicolumn{1}{c|}{34.3} & 12.30 & 7.11 & 45 & 35 & 28 \\
				\multicolumn{1}{c|}{} & NaVid~\cite{zhang2024navid}& - & - & - & - & \multicolumn{1}{c|}{-} & - & \textbf{5.47} & 49.1 & 37.4 & \multicolumn{1}{c|}{\textbf{35.9}} & - & - & - & - & - \\
				\multicolumn{1}{c|}{} & ETPNav/p~\cite{wang2024sim}  & - & - & - & - & \multicolumn{1}{c|}{-} & - & 6.81 & 42.4 & 32.9 & \multicolumn{1}{c|}{23.1} & - & - & - & - & - \\ \cmidrule(l){2-17} 
				\multicolumn{1}{c|}{} & VLN-3DFF~\cite{wang2024sim} & - & - & - & - & \multicolumn{1}{c|}{-} & - & 5.95 & 55.8 & 44.9 & \multicolumn{1}{c|}{30.4} & - & 6.24 & 54.4 & 43.7 & 28.9 \\
				\multicolumn{1}{c|}{} & \cellcolor{gray!10}VLN-3DFF* & \cellcolor{gray!10}22.90 & \cellcolor{gray!10}4.92 & \cellcolor{gray!10}62.1 & \cellcolor{gray!10}52.7 & \multicolumn{1}{c|}{\cellcolor{gray!10}36.7} & \cellcolor{gray!10}26.16 & \cellcolor{gray!10}6.05 & \cellcolor{gray!10}54.9 & \cellcolor{gray!10}43.8 & \multicolumn{1}{c|}{\cellcolor{gray!10}29.4} & \cellcolor{gray!10}26.02 & \cellcolor{gray!10}6.22 & \cellcolor{gray!10}{\textbf{54.7}} & \cellcolor{gray!10}43.8 & \cellcolor{gray!10}28.6 \\
				\multicolumn{1}{c|}{\multirow{-8}{*}{\textbf{Monocular}}} & \cellcolor{gray!25}\textbf{NavMorph} & \cellcolor{gray!25}\textcolor{blue}{20.03} & \cellcolor{gray!25}\textbf{\textcolor{blue}{4.58}} & \cellcolor{gray!25}\textbf{\textcolor{blue}{62.7}} & \cellcolor{gray!25}\textbf{\textcolor{blue}{55.8}} & \multicolumn{1}{c|}{\cellcolor{gray!25}{\textcolor{blue}{38.9}}} & \cellcolor{gray!25}\textcolor{blue}{22.54} & \cellcolor{gray!25}\textcolor{blue}{5.75} & \cellcolor{gray!25}\textbf{\textcolor{blue}{56.9}} & \cellcolor{gray!25}\textbf{\textcolor{blue}{47.9}} & \multicolumn{1}{c|}{\cellcolor{gray!25}\textcolor{blue}{33.2}} & \cellcolor{gray!25}\textcolor{blue}{24.75} & \cellcolor{gray!25}\textbf{\textcolor{blue}{6.01}} & \cellcolor{gray!25}54.5 & \cellcolor{gray!25}{\textcolor{blue}{\textbf{45.7}}} & \cellcolor{gray!25}{\textcolor{blue}{\textbf{30.2}}} \\ \midrule[0.7pt]
				\multicolumn{1}{c|}{} 
				& Seq2Seq~\cite{anderson2018vision} & {9.26} & 7.12 & 46 & 37 & \multicolumn{1}{c|}{35} & {8.64} & 7.37 & 40 & 32 & \multicolumn{1}{c|}{30} & {8.85} & 7.91 & 36 & 28 & 25 \\
				& SASRA~\cite{Irshad2021SemanticallyawareSR}  
				& 8.89  & 7.71  & -  & 36  &  \multicolumn{1}{c|}{34}
				& 7.89  & 8.32 & - & 24 & \multicolumn{1}{c|}{22} 
				& -  & - & - & - & -\\
				& CWTP~\cite{Chen2020TopologicalPW} 
				& -  & 7.10  & 56  & 36  & \multicolumn{1}{c|}{31}  
				& -  & 7.90 & 38 & 26 & \multicolumn{1}{c|}{23} 
				& -  & - & - & - & - \\
				& AG-CMTP~\cite{Chen2022ReinforcedSS}  
				& - & 6.60 & 56 & 36 & \multicolumn{1}{c|}{31} 
				& - & 7.90 & 39 & 23 & \multicolumn{1}{c|}{19}
				& -  & - & - & - & - \\
				& R2R-CMTP~\cite{Chen2022ReinforcedSS}  
				& - & 7.10 & 45 & 36 & \multicolumn{1}{c|}{31} 
				& - & 7.90 & 38 & 26 & \multicolumn{1}{c|}{23} 
				& - & - & - & - & -  \\
				& WPN~\cite{wang2024Graphbe}  
				& \textbf{8.54} & 5.48 & 53 & 46 & \multicolumn{1}{c|}{43}
				& \textbf{7.62} & 6.31 & 40 & 36 & \multicolumn{1}{c|}{34} 
				& \textbf{8.02} & 6.65 & 37 & 32 & 30 \\ 
				& Sim2Sim~\cite{Krantz2022Sim2SimTF}& 11.18 & 4.67 & 61 & 52 & \multicolumn{1}{c|}{44} & 10.69 & 6.07 & 52 & 43 & \multicolumn{1}{c|}{36} & 11.43 & 6.17 & 52 & 44 & 37 \\
				& CWP-CMA~\cite{Hong2022BridgingTG}& 11.47 & 5.20 & 61 & 51 & \multicolumn{1}{c|}{45} & 10.90 & 6.20 & 52 & 41 & \multicolumn{1}{c|}{36} & 11.85 & 6.30 & 49 & 38 & 33 \\
				& CWP-BERT~\cite{Hong2022BridgingTG}& 12.50 & 5.02 & 59 & 50 & \multicolumn{1}{c|}{44} & 12.23 & 5.74 & 53 & 44 & \multicolumn{1}{c|}{39} & 13.51 & 5.89 & 51 & 42 & 36 \\
				& ERG~\cite{wang2024Graphbe}  
				& 11.80 & 5.04 &61 & 46 & \multicolumn{1}{c|}{42}
				& 9.96 & 6.20 & 52 & 41 & \multicolumn{1}{c|}{36}
				& - & - & - & - & -  \\   
				& DUET~\cite{chen2022think}  
				& 12.62 & 4.13 & 67 & 57 & \multicolumn{1}{c|}{49}
				& 11.86 & 5.13 & 55 & 46 & \multicolumn{1}{c|}{40} 
				& 13.13 & 5.82 & 50 & 42 & 36\\
				& DREAMW~\cite{Wang2023DREAMWALKERMP} & 11.60 & 4.09 & 59 & 66 & \multicolumn{1}{c|}{48} & 11.30 & 5.53 & 49 & {59} & \multicolumn{1}{c|}{44} & 11.80 & 5.48 & 49 & {57} & 44 \\
				& Ego$^2$-Map~\cite{Hong2023LearningNV} & - & - & - & - & \multicolumn{1}{c|}{-} & - & 4.93 & - & {52} & \multicolumn{1}{c|}{46} & - & 5.54 & 56 & {47} & 41 \\
				& ScaleVLN~\cite{wang2023scaling} & - & - & - & - & \multicolumn{1}{c|}{-} & - & 4.80 & - & {55} & \multicolumn{1}{c|}{51} & - & 5.11 & - & {55} & 50 \\
				& GridMM~\cite{wang2023gridmm}& 12.69 & 4.21 & 69 & 59 & \multicolumn{1}{c|}{51} & 13.36 & 5.11 & 61 & 49 & \multicolumn{1}{c|}{41} & 13.31 & 5.64 & 56 & 46 & 39 \\
				& BEVBert~\cite{an2023bevbert} & 13.98 & 3.77 & 73 & 68 & \multicolumn{1}{c|}{60} & 13.27 & 4.57 & 67 & 59 & \multicolumn{1}{c|}{50} & 15.31 & 4.70 & 67 & 59 & 50 \\ 
				%& HNR~\cite{wang2024lookahead} & 11.79 & 3.67 & 76 & 69 & \multicolumn{1}{c|}{61} &12.64 & 4.42 & 67 & 61 & \multicolumn{1}{c|}{51} & 13.03 & 4.81 & 67 & 58 & 50\\
				& FSTTA~\cite{gao2024fast} & 12.39 & 4.25 & 69 & 58 & \multicolumn{1}{c|}{50} & 11.58 & 5.27 & 58 & 48 & \multicolumn{1}{c|}{42} & 13.17 & 5.84 & 55 & 46 & 38 \\ \cmidrule(l){2-17} 
				%& UnitedVLN~\cite{Dai2024UnitedVLNGG} & - & \textbf{3.30} & \textbf{78} & \textbf{70} & \multicolumn{1}{c|}{61} & - & \textbf{4.26} & \textbf{70} & 62 & \multicolumn{1}{c|}{49} & - & \textbf{4.67} & \textbf{68} & 57 & 47 \\ \cmidrule(l){2-17} 
				& ETPNav~\cite{an2024etpnav} & 11.78 & 3.95 & 72 & 66 & \multicolumn{1}{c|}{59} & 11.99 & 4.71 & 65 & 57 & \multicolumn{1}{c|}{49} & 12.87 & 5.12 & 63 & 55 & 48 \\ 
				& \cellcolor{gray!10}ETPNav* & \multicolumn{1}{c}{\cellcolor{gray!10}11.35} & \multicolumn{1}{c}{\cellcolor{gray!10}3.93} & \multicolumn{1}{c}{\cellcolor{gray!10}72} & \multicolumn{1}{c}{\cellcolor{gray!10}66} &\multicolumn{1}{c|}{\cellcolor{gray!10}59} & \multicolumn{1}{c}{\cellcolor{gray!10}11.40} & \multicolumn{1}{c}{\cellcolor{gray!10}4.69} & \multicolumn{1}{c}{\cellcolor{gray!10}64} & \multicolumn{1}{c}{\cellcolor{gray!10}57} & \multicolumn{1}{c|}{\cellcolor{gray!10}49} & \cellcolor{gray!10}12.72 & \cellcolor{gray!10}5.10 & \cellcolor{gray!10}63 & \cellcolor{gray!10}55 & \cellcolor{gray!10}48 \\
				& \cellcolor{gray!25}\textbf{NavMorph} & \multicolumn{1}{c}{\cellcolor{gray!25}11.43} &  \multicolumn{1}{c}{\cellcolor{gray!25}{\textcolor{blue}{3.86}}} &  \multicolumn{1}{c}{\cellcolor{gray!25}{\textcolor{blue}{73}}} &  \multicolumn{1}{c}{\cellcolor{gray!25}{\textcolor{blue}{67}}} &                          \multicolumn{1}{c|}{\cellcolor{gray!25}{\textcolor{blue}{60}}} & \multicolumn{1}{c}{\cellcolor{gray!25}11.55} & \multicolumn{1}{c}{\cellcolor{gray!25}{\textcolor{blue}{4.62}}} & \multicolumn{1}{c}{\cellcolor{gray!25}{\textcolor{blue}{66}}} & \multicolumn{1}{c}{\cellcolor{gray!25}{\textcolor{blue}{59}}} & \multicolumn{1}{c|}{\cellcolor{gray!25}{\textcolor{blue}{50}}} & \cellcolor{gray!25}12.88 & \cellcolor{gray!25}{\textcolor{blue}{4.91}} & \cellcolor{gray!25}{\textcolor{blue}{64}} & \cellcolor{gray!25}{\textcolor{blue}{57}} & \cellcolor{gray!25}{\textcolor{blue}{49}}\\\cmidrule(l){2-17} 
				& HNR~\cite{wang2024lookahead} & 11.79 & 3.67 & 76 & 69 & \multicolumn{1}{c|}{61} & 12.64 & 4.42 & 67 & 61 & \multicolumn{1}{c|}{51} & 13.03 & 4.81 & 67 & 58 & 50 \\ 
				& \cellcolor{gray!10}HNR* & \multicolumn{1}{c}{\cellcolor{gray!10}11.84} & \multicolumn{1}{c}{\cellcolor{gray!10}3.73} & \multicolumn{1}{c}{\cellcolor{gray!10}76} & \multicolumn{1}{c}{\cellcolor{gray!10}69} &\multicolumn{1}{c|}{\cellcolor{gray!10}61} & \multicolumn{1}{c}{\cellcolor{gray!10}12.76} & \multicolumn{1}{c}{\cellcolor{gray!10}4.57} & \multicolumn{1}{c}{\cellcolor{gray!10}67} & \multicolumn{1}{c}{\cellcolor{gray!10}61} & \multicolumn{1}{c|}{\cellcolor{gray!10}51} & \cellcolor{gray!10}12.92 & \cellcolor{gray!10}4.85 & \cellcolor{gray!10}67 & \cellcolor{gray!10}58 & \cellcolor{gray!10}50 \\
				\multirow{-24}{*}{\textbf{Panoramic}} & \cellcolor{gray!25}\textbf{NavMorph} & \multicolumn{1}{c}{\cellcolor{gray!25}\textcolor{blue}{11.76}} &  \multicolumn{1}{c}{\cellcolor{gray!25}\textbf{\textcolor{blue}{3.66}}} &  \multicolumn{1}{c}{\cellcolor{gray!25}\textbf{\textcolor{blue}{78}}} &  \multicolumn{1}{c}{\cellcolor{gray!25}\textbf{\textcolor{blue}{70}}} &  \multicolumn{1}{c|}{\cellcolor{gray!25}\textbf{\textcolor{blue}{62}}} & \multicolumn{1}{c}{\cellcolor{gray!25}\textcolor{blue}{12.68}} & \multicolumn{1}{c}{\cellcolor{gray!25}\textbf{\textcolor{blue}{4.37}}} & \multicolumn{1}{c}{\cellcolor{gray!25}\textbf{\textcolor{blue}{68}}} & \multicolumn{1}{c}{\cellcolor{gray!25}\textbf{\textcolor{blue}{64}}} & \multicolumn{1}{c|}{\cellcolor{gray!25}\textbf{\textcolor{blue}{53}}} & \cellcolor{gray!25}\textcolor{blue}{12.69} & \cellcolor{gray!25}\textbf{\textcolor{blue}{4.69}} & \cellcolor{gray!25}\textbf{\textcolor{blue}{68}} & \cellcolor{gray!25}\textbf{\textcolor{blue}{60}} & \cellcolor{gray!25}\textbf{\textcolor{blue}{52}}\\
				\bottomrule
			\end{tabular}
			\begin{tablenotes}    
				\footnotesize                              
				\item Note: Following established conventions in prior works, we report experimental results with different precision formats across camera configurations: integers for panoramic settings and two decimal places for monocular settings.
			\end{tablenotes}           
		\end{threeparttable} 	
	}
\end{table*}

\section{Complementary Experiments}\label{sec:exper}
\subsection{Full Results}
In our main paper, we provide representative comparison results on the R2R-CE~\cite{krantz2020beyond,anderson2018vision} and RxR-CE~\cite{krantz2020beyond,ku2020room} benchmarks due to space constraints. Here, we present the complete results across the `validation seen', `validation unseen', and `test unseen' splits of these benchmarks, including comparisons with a broader range of state-of-the-art methods, as detailed in Table~\ref{tab:r2r-ce-full} and Table~\ref{tab:rxr-ce-full}. Our self-evolving world model enhances its ability to anticipate future states based on current observations and cross-episodic experiences, effectively handling complex navigation tasks even with monocular input.

\noindent\textbf{Performance Improvement on Seen/Unseen sets.}
Based on the experimental results in Table~\ref{tab:r2r-ce-full} and Table~\ref{tab:rxr-ce-full}, our proposed NavMorph consistently achieves notable performance improvements across different datasets. While performance gains varies between seen and unseen environments, we analyze relative improvements to better quantify the effectiveness of our self-evolving world model across different settings. 

Taking the monocular setup as an example, NavMorph improves the success rate (SR) by 6.85$\%$ in unseen environments, compared to 5.88$\%$ in seen environments on R2R-CE dataset. 
The improvement in SPL is even more pronounced, reaching 9.26$\%$ in unseen environments versus 5.99$\%$ in seen environments. A similar trend is observed in RxR-CE, where unseen SR improves by 10.94$\%$, while seen SR increases by 7.54$\%$. Likewise, SPL improves 11.29$\%$ in unseen settings, compared to 12.71$\%$ in seen ones.
These results indicate that NavMorph achieves higher or comparable performance gains in unseen environments (average of val/test unseen) compared to seen ones, demonstrating its capacity to generalize across novel tasks.

A key factor contributing to this generalization ability is self-evolution, which enhances adaptation uniformly across both seen and unseen data rather than specifically optimizing for new scenarios. The observed gains in seen settings further suggest that the model effectively adapts to novel instructions within familiar scenes, rather than merely overfitting to training data. 

\begin{table*}[ht] \centering
	\caption{Experimental results on RxR-CE datasets. Results better than the base model are shown in blue. Best results for the panoramic and monocular settings are each highlighted in bold.}
	\vspace{-3mm}
	\label{tab:rxr-ce-full}
	\resizebox{\textwidth}{!}{
		\begin{threeparttable}
			\begin{tabular}{@{}c|l|ccccccc|ccccccc|ccccccc@{}}
				\toprule
				& \multicolumn{1}{c|}{} & \multicolumn{7}{c|}{\textbf{Val Seen}} & \multicolumn{7}{c|}{\textbf{Val Unseen}} & \multicolumn{7}{c}{\textbf{Test Unseen}} \\
				\multirow{-2}{*}{\textbf{Camera}} & \multicolumn{1}{c|}{\multirow{-2}{*}{\textbf{Methods}}} &		
				\cellcolor{red!25}TL $\downarrow$& \cellcolor{red!25}NE $\downarrow$& \cellcolor{gray!25}OSR & \cellcolor{gray!25}SR & \cellcolor{gray!25}SPL & \cellcolor{gray!25}NDTW & \cellcolor{gray!25}SDTW & \cellcolor{red!25}TL $\downarrow$& \cellcolor{red!25}NE $\downarrow$& \cellcolor{gray!25}OSR & \cellcolor{gray!25}SR & \cellcolor{gray!25}SPL & \cellcolor{gray!25}NDTW & \cellcolor{gray!25}SDTW & \cellcolor{red!25}TL $\downarrow$& \cellcolor{red!25}NE $\downarrow$& \cellcolor{gray!25}OSR & \cellcolor{gray!25}SR & \cellcolor{gray!25}SPL & \cellcolor{gray!25}NDTW & \cellcolor{gray!25}SDTW \\ \midrule
				\multicolumn{1}{l|}{} & LAW~\cite{Raychaudhuri2021LanguageAlignedW} & \textbf{7.92} & 11.94 & 20.0 & 7.0 & 6.0 & - & - & \textbf{4.01} & 10.87 & 21.0 & 8.0 & 8.0 & - & - & - & - & - & - & - & - & - \\
				\multicolumn{1}{l|}{} & CM$^2$~\cite{Georgakis2022CrossmodalML} & - & - & - & - & - & - & - & 12.29 & 8.98 & 25.3 & 14.4 & 9.2 & - & - & - & - & - & - & - & - & - \\
				\multicolumn{1}{l|}{} & WS-MGMap~\cite{Chen2022WeaklySupervisedMM} & 10.37 & 10.19 & 27.7 & 14.0 & 12.3 & - & - & 10.80 & 9.83 & 29.8 & 15.0 & 12.1 & - & - & - & - & - & - & - & - & - \\
				\multicolumn{1}{l|}{} & NaVid~\cite{zhang2024navid} & - & - & - & - & - & - & - & 10.59 & \textbf{8.41} & 34.5 & 23.8 & 32.2 & - & - & - & - & - & - & - & - & - \\
				\multicolumn{1}{l|}{} & A$^2$-Nav~\cite{chen20232} & - & - & - & - & - & - & - & - & - & - & 16.8 & 6.3 & - & - & - & - & - & - & - & - & - \\ \cmidrule(l){2-23} 
				\multicolumn{1}{l|}{} & VLN-3DFF~\cite{wang2024sim} & - & - & - & - & - & - & - & - & 8.79 & 36.7 & 25.5 & 18.1 & - & - & - & - & - & - & - & - & - \\
				\multicolumn{1}{l|}{} & \cellcolor{gray!10}VLN-3DFF* & \cellcolor{gray!10}18.91 & \cellcolor{gray!10}9.87 & \cellcolor{gray!10}40.54 & \cellcolor{gray!10}27.72 & \cellcolor{gray!10}20.61 & \cellcolor{gray!10}42.37 & \cellcolor{gray!10}20.94 & \cellcolor{gray!10}16.21 & \cellcolor{gray!10}9.41 & \cellcolor{gray!10}38.40 & \cellcolor{gray!10}26.66 & \cellcolor{gray!10}20.11 & \cellcolor{gray!10}42.91 & \cellcolor{gray!10}20.36 & \cellcolor{gray!10}\textbf{20.85}		
				& \cellcolor{gray!10}10.19 & \cellcolor{gray!10}- & \cellcolor{gray!10}23.41 & \cellcolor{gray!10}15.43 & \cellcolor{gray!10}32.38 & \cellcolor{gray!10}14.75 \\
				\multicolumn{1}{l|}{\multirow{-8}{*}{\textbf{Monocular}}} & \cellcolor{gray!25}\textbf{NavMorph} & \cellcolor{gray!25}21.61 & \cellcolor{gray!25}\textcolor{blue}{\textbf{9.80}} & \cellcolor{gray!25}\textcolor{blue}{\textbf{41.27}} & \cellcolor{gray!25}\textcolor{blue}{\textbf{29.81}} & \cellcolor{gray!25}\textcolor{blue}{\textbf{23.23}} & \cellcolor{gray!25}\textcolor{blue}{\textbf{44.51}} & \cellcolor{gray!25}\textcolor{blue}{\textbf{22.68}} & \cellcolor{gray!25}{20.28} & \cellcolor{gray!25}\textcolor{blue}{8.85} & \cellcolor{gray!25}\textcolor{blue}{\textbf{43.05}} & \cellcolor{gray!25}\textcolor{blue}{\textbf{30.76}} & \cellcolor{gray!25}\textcolor{blue}{\textbf{22.84}} & \cellcolor{gray!25}\textcolor{blue}{\textbf{44.19}} & \cellcolor{gray!25}\textcolor{blue}{\textbf{23.30}} & \cellcolor{gray!25}21.13			
				& \cellcolor{gray!25}\textcolor{blue}{\textbf{9.81}}& \cellcolor{gray!25}- & \cellcolor{gray!25}\textcolor{blue}{\textbf{24.93}} & \cellcolor{gray!25}\textcolor{blue}{\textbf{16.82}} & \cellcolor{gray!25}\textcolor{blue}{\textbf{33.71}}  & \cellcolor{gray!25}\textcolor{blue}{\textbf{15.64}} \\ \midrule[0.7pt]
				& Seq2Seq~\cite{anderson2018vision} & - & - & - & - & - & - & - & 7.33 & 12.1 & - & 13.93 & 11.96 & 30.86 & 11.01 & - & 12.10 & - & 13.93 & 11.96 & 30.86 & 11.01 \\
				& Reborn~\cite{an20221st} & - & 5.69 & - & 52.43 & 45.46 & 66.27 & 44.47 & - & 5.98 & - & 48.60 & 42.05 & 63.35 & 41.82 & - & 7.10 & - & 45.82 & 38.82 & 55.43 & 38.42 \\
				& CWP-CMA~\cite{Hong2022BridgingTG} & - & - & - & - & - & - & - & - & 8.76 & - & 26.59 & 22.16 & 47.05 & - & \textbf{20.04} & 10.4 & - & 24.08 & 19.07 & 37.39 & 18.65 \\
				& CWP-RecBERT~\cite{Hong2022BridgingTG} & - & - & - & - & - & - & - & - & 8.98 & - & 27.08 & 22.65 & 46.71 & - & 20.09 & 10.4 & - & 24.85 & 19.61 & 37.30 & 19.05 \\
				& AO-Planner~\cite{chen2024affordances} & - & - & - & - & - & - & - & - & 7.06 & - & 43.3 & 30.5 & 50.1 & - & - & - & - & - & - & - & - \\
				& LAW-Pano~\cite{Raychaudhuri2021LanguageAlignedW} & \textbf{6.27} & 12.07 & 17.0 & 9.0 & 9.0 & - & - & \textbf{4.62} & 11.04 & 16.0 & 10.0 & 9.0 & - & - & - & - & - & - & - & - & - \\
				& UnitedVLN~\cite{Dai2024UnitedVLNGG}& - & \textbf{4.74} & - & \textbf{65.1} & 52.9 & 69.4 & 53.6 & - & \textbf{5.48} & - & 57.9 & 45.9 & 63.9 & 48.1 & - & - & - & - & - & - & - \\ \cmidrule(l){2-23} 
				& ETPNav~\cite{an2024etpnav} & - & 5.03 & - & 61.46 & 50.83 & 66.41 & 51.28 & - & 5.64 & - & 54.79 & 44.89 & 61.90 & 45.33 & - & 6.99 & - & 51.21 & 39.86 & 54.11 & 41.30 \\
				& \cellcolor{gray!10}ETPNav* & \cellcolor{gray!10}18.16				
				& \cellcolor{gray!10}5.06 & \cellcolor{gray!10}64.06 & \cellcolor{gray!10}62.09 & \cellcolor{gray!10}50.64 & \cellcolor{gray!10}66.06 & \cellcolor{gray!10}51.17 & \cellcolor{gray!10}{18.92} & \cellcolor{gray!10}{5.96} & \cellcolor{gray!10}{63.66} & \cellcolor{gray!10}54.83 & \cellcolor{gray!10}44.62 & \cellcolor{gray!10}61.36 & \cellcolor{gray!10}44.87 &  \cellcolor{gray!10}21.83 & \cellcolor{gray!10}6.92 & \cellcolor{gray!10}- & \cellcolor{gray!10}51.38 & \cellcolor{gray!10}39.90 & \cellcolor{gray!10}53.85 & \cellcolor{gray!10}40.91 \\
				& \cellcolor{gray!25}\textbf{NavMorph} & \cellcolor{gray!25}{18.97} & \cellcolor{gray!25}{5.08} & \cellcolor{gray!25}\textcolor{blue}{65.86} & \cellcolor{gray!25}\textcolor{blue}{63.88} & \cellcolor{gray!25}\textcolor{blue}{52.28} & \cellcolor{gray!25}\textcolor{blue}{67.94} & \cellcolor{gray!25}\textcolor{blue}{52.54} & \cellcolor{gray!25}{19.93} & \cellcolor{gray!25}\textcolor{blue}{5.80} & \cellcolor{gray!25}\textcolor{blue}{64.83} & \cellcolor{gray!25}\textcolor{blue}{56.23} & \cellcolor{gray!25}\textcolor{blue}{46.39} & \cellcolor{gray!25}\textcolor{blue}{63.23} & \cellcolor{gray!25}\textcolor{blue}{46.98} & \cellcolor{gray!25}\textcolor{blue}{21.29} & \cellcolor{gray!25}\textcolor{blue}{6.90}	& \cellcolor{gray!25}- & \cellcolor{gray!25}\textcolor{blue}{51.97} & \cellcolor{gray!25}\textcolor{blue}{41.56} & \cellcolor{gray!25}\textcolor{blue}{55.01} & \cellcolor{gray!25}\textcolor{blue}{42.60} \\ \cmidrule(l){2-23} 
				& HNR~\cite{wang2024lookahead} & - & {4.85} & - & 63.72 & 53.17 & 68.81 & 52.78 & - & 5.51 & - & 56.39 & 46.73 & 63.56 & 47.24 & - & 6.81 & - & 53.22 & 41.14 & 55.61 & 42.89 \\
				& \cellcolor{gray!10}HNR* & \cellcolor{gray!10}19.74 & \cellcolor{gray!10}4.93 & \cellcolor{gray!10}66.01 & \cellcolor{gray!10}63.55 & \cellcolor{gray!10}53.37 & \cellcolor{gray!10}69.02 & \cellcolor{gray!10}52.66  & \cellcolor{gray!10}20.41 & \cellcolor{gray!10}5.75 & \cellcolor{gray!10}64.93  & \cellcolor{gray!10}56.48 & \cellcolor{gray!10}46.62 & \cellcolor{gray!10}63.43 & \cellcolor{gray!10}47.38 & \cellcolor{gray!10}23.02&\cellcolor{gray!10}6.88 & \cellcolor{gray!10}- & \cellcolor{gray!10}53.33 & \cellcolor{gray!10}41.18 & \cellcolor{gray!10}55.47 & \cellcolor{gray!10}42.95 \\
				\multirow{-14}{*}{\textbf{Panoramic}} & \cellcolor{gray!25}\textbf{NavMorph} & \cellcolor{gray!25}{20.80} & \cellcolor{gray!25}{5.10} & \cellcolor{gray!25}\textcolor{blue}{\textbf{67.88}} & \cellcolor{gray!25}\textcolor{blue}{{64.95}} & \cellcolor{gray!25}\textcolor{blue}{\textbf{54.17}} & \cellcolor{gray!25}\textcolor{blue}{\textbf{70.94}} & \cellcolor{gray!25}\textcolor{blue}{\textbf{54.82}} & \cellcolor{gray!25}{21.33} & \cellcolor{gray!25}\textcolor{blue}{{5.67}} &\cellcolor{gray!25}\textcolor{blue}{\textbf{66.02}}  & \cellcolor{gray!25}\textcolor{blue}{\textbf{58.02}}  & \cellcolor{gray!25}\textcolor{blue}{\textbf{48.98}}  & \cellcolor{gray!25}\textcolor{blue}{\textbf{64.77}}  & \cellcolor{gray!25}\textcolor{blue}{\textbf{48.85}}& \cellcolor{gray!25}23.36 & \cellcolor{gray!25}\textcolor{blue}{\textbf{6.67}}	& \cellcolor{gray!25}- & \cellcolor{gray!25}\textcolor{blue}{\textbf{54.98}} & \cellcolor{gray!25}\textcolor{blue}{\textbf{43.02}} & \cellcolor{gray!25}\textcolor{blue}{\textbf{57.31}} & \cellcolor{gray!25}\textcolor{blue}{\textbf{44.76}} \\
				\bottomrule
			\end{tabular}
			\begin{tablenotes}    
				\footnotesize                              
				\item Note: Official evaluation on the Test Unseen split of RxR-CE dataset only provides TL, NE, SR, SPL, NDTW and SDTW metrics, thus OSR metric is not reported for the test split in this table.
			\end{tablenotes}           
		\end{threeparttable} 	
	}
\end{table*}

\subsection{Extended Results for Self-Evolution}
\noindent\textbf{Detailed Ablation Study on Self-Evolution Strategy.}
In our main paper (\underline{\textit{Table 3}}), we conducted an ablation study on the effect of self-evolution, in which the proposed Contextual Evolution Memory (CEM) module was entirely prevented from updating. The results demonstrated the effectiveness of self-evolution in enhancing model performance and learning dynamics.
To further investigate its role only in online adaptation, we introduce `NavMorph \textit{w/o} SE*', a variant where CEM undergoes self-evolution following \underline{\textit{Eq. (3) in main paper}} during training, progressively refining its stored representations. Once training is complete, the finalized memory is used as the initial state for deployment and remains unchanged throughout online testing.

%As shown in Table~\ref{tab:ablation-se}, even after training with self-evolution, disabling it during testing leads to performance degradation in online unseen environments, reinforcing its importance for real-time adaptation.
As shown in Table~\ref{tab:ablation-se}, enabling self-evolution during online testing improves performance in online unseen environments, highlighting its crucial role in real-time adaptation.
Moreover, since the self-evolution process benefits from prolonged environmental interaction—where unsupervised learning progressively refines the model’s dynamic latent state—we extend our analysis to a larger, more diverse dataset, RxR-CE, to examine its influence on generalization. The results indicate a notable improvement in SPL (21.46$\to$22.84), further validating the effectiveness of self-evolution in enhancing adaptability in unseen environments.

\noindent\textbf{Different Steps of Predictive Future States $T_p$.}
As shown in Table~\ref{tab:steps}, we further investigate the impact of varying predictive steps in our foresight action planner on navigational performance. Notably, predicting two steps ($T_p = 2$) achieves the optimal balance across key metrics, offering sufficient foresight for reliable decision-making without introducing excessive uncertainty.
As $T_p$ increases beyond 2, we observe a slight decline in SPL, SR, and NDTW, possibly due to compounding errors (accumulation of inaccuracies over multiple predictive steps) or over-commitment to future predictions (focuses too heavily on long-term predictions), which reduces the agent's flexibility to adapt to changing environments. 
These results demonstrate the need of striking a balance between foresight and adaptability. Predicting too few steps may limit the agent’s strategic planning, while predicting too many steps introduces unnecessary complexity, diminishing trajectory efficiency.

\begin{table}\centering
	\caption{Ablation Study on Self-Evolution.}
	\label{tab:ablation-se}
	\resizebox{0.98\linewidth}{!}{
		\begin{threeparttable} 
			\begin{tabular}{@{}c|c|ccccccc@{}}
				\toprule
				\textbf{Dataset} & \textbf{Methods} & \cellcolor{red!25}TL $\downarrow$ & \cellcolor{red!25}NE $\downarrow$ & \cellcolor{gray!25}OSR & \cellcolor{gray!25}SR & \cellcolor{gray!25}SPL & \cellcolor{gray!25}NDTW & \cellcolor{gray!25}SDTW \\ \midrule
				\multirow{3}{*}{\begin{tabular}[c]{@{}c@{}}R2R-CE\\ Val\\ Unseen \end{tabular}} & Base Model & 26.16 & 6.05 & 54.92 & 43.77 & 29.39 & 40.94 & 29.30 \\\cmidrule{2-9}
				& \cellcolor{gray!0}{{NavMorph \textit{w/o} SE}*} & \cellcolor{gray!0}{23.33} & \cellcolor{gray!0}{5.77} & \cellcolor{gray!0}{56.12} & \cellcolor{gray!0}{46.87} & \cellcolor{gray!0}{32.56} & \cellcolor{gray!0}{44.42} & \cellcolor{gray!0}{32.16} \\
				& \cellcolor{gray!25}{\textbf{NavMorph}} & \cellcolor{gray!25}{\textcolor{blue}{22.54}} & \cellcolor{gray!25}{\textcolor{blue}{5.75}} & \cellcolor{gray!25}{\textcolor{blue}{56.88}} & \cellcolor{gray!25}{\textcolor{blue}{47.91}} & \cellcolor{gray!25}{\textcolor{blue}{33.22}} & \cellcolor{gray!25}{\textcolor{blue}{44.86}} &\cellcolor{gray!25}{\textcolor{blue}{32.73}} \\ \midrule
				\multirow{3}{*}{\begin{tabular}[c]{@{}c@{}}RxR-CE\\ Val\\ Unseen\end{tabular}} & Base Model & 16.21 & 9.41 & 38.40 & 26.66 & 20.11 & 42.91 & 20.36 \\\cmidrule{2-9}
				& \cellcolor{gray!0}{{NavMorph \textit{w/o} SE}*} & \cellcolor{gray!0}{20.83} & \cellcolor{gray!0}{9.08} & \cellcolor{gray!0}{41.49} & \cellcolor{gray!0}{28.78} & \cellcolor{gray!0}{21.46} & \cellcolor{gray!0}{43.26} & \cellcolor{gray!0}{21.52}\\
				& \cellcolor{gray!25}{\textbf{NavMorph}} & \cellcolor{gray!25}\textcolor{blue}{20.28} & \cellcolor{gray!25}{\textcolor{blue}{8.85}} & \cellcolor{gray!25}{\textcolor{blue}{43.05}} & \cellcolor{gray!25}{\textcolor{blue}{30.76}} & \cellcolor{gray!25}{\textcolor{blue}{22.84}} & \cellcolor{gray!25}{\textcolor{blue}{44.19}} & \cellcolor{gray!25}{\textcolor{blue}{23.30}} \\ \bottomrule
			\end{tabular}
			\begin{tablenotes}    
				\footnotesize               
				\item Note: Results better than `NavMorph \textit{w/o} SE*' are shown in blue.
			\end{tablenotes}           
		\end{threeparttable} 
	}
\end{table}

\begin{table} \centering
	\caption{Experimental Results for Different Predictive Steps.}
	\vspace{-3mm}
	\label{tab:steps}
	\resizebox{0.98\linewidth}{!}{
		\begin{tabular}{@{}c|c|ccccccc@{}}
			\toprule
			\multirow{2}{*}{\textbf{Methods}} & \multirow{2}{*}{\textbf{\begin{tabular}[c]{@{}c@{}}Predictive\\ Steps $T_p$\end{tabular}}} & \multicolumn{7}{c}{\textbf{R2R-CE Val Unseen}} \\
			&  & \cellcolor{red!25}TL $\downarrow$ & \cellcolor{red!25}NE $\downarrow$ & \cellcolor{gray!25}OSR & \cellcolor{gray!25}SR & \cellcolor{gray!25}SPL & \cellcolor{gray!25}NDTW & \cellcolor{gray!25}SDTW  \\ \midrule
			\multicolumn{1}{c|}{Base model} & -  & 26.16 & 6.05 & 54.92 & 43.77 & 29.39 & 40.94 & 29.30 \\ \midrule
			%\multirow{5}{*}{\begin{tabular}[c]{@{}c@{}}Our\\ World\\ Model\end{tabular}} & 1 & 22.05 & 5.99 & 55.57 & 46.06 & 32.78 & \textbf{44.89} & 32.36 \\
			\multirow{5}{*}{NavMorph} & 1 & 22.05 & 5.99 & 55.57 & 46.06 & 32.78 & \textbf{44.89} & 32.36 \\
			&\cellcolor{gray!25}\textbf{2} & \cellcolor{gray!25}22.54 & \cellcolor{gray!25}5.75 & \cellcolor{gray!25}\textbf{56.88} & \cellcolor{gray!25}\textbf{47.91} & \cellcolor{gray!25}\textbf{33.22} & \cellcolor{gray!25}44.86 & \cellcolor{gray!25}\textbf{32.73} \\
			& 3 & 25.36 & 5.99 & 56.50 & 44.97 & 31.30 & 42.99 & 30.57 \\
			& 4 & \textbf{20.91} & 5.81 & 55.52 & 46.82 & 32.04 & 44.61 & 32.20 \\
			& 5 & 25.94 & \textbf{5.69} & 56.66 & 47.18 & 31.79 & 43.72 & 31.92 \\ \bottomrule
		\end{tabular}
	}
	\vspace{-4mm}
\end{table}

\subsection{Other Ablation Studies}
\noindent\textbf{Comparison with Representative TTA Strategies.}
In our main paper, we discussed how our evolving world model accumulates scene-specific information from test environments as memory knowledge during online testing. This mechanism allows the model to refine its predictions in dynamically changing environments without requiring ground truth actions.
A related class of approaches, referred to as Test-Time Adaptation (TTA), also aims to improve model generalization by dynamically adjusting model parameters during testing, often through gradient-based updates or statistical alignment methods (\textit{e.g.}, batch normalization adaptation, entropy minimization).

Given the comparison with the most related method, FSTTA~\cite{gao2024fast}, in our main paper, we further evaluate representative TTA approaches under the same test conditions as FSTTA (\textit{i.e.,} updating the same set of parameters) and compare them with our NavMorph. To ensure a fair and robust evaluation, we conduct experiments under three different random seeds, reporting both the mean and standard deviation of the results.
As demonstrated in Table~\ref{tab:ablation-tta}, our proposed NavMorph achieves the best overall performance while exhibiting more stable results (with lower standard deviation). These results highlight the effectiveness of incorporating a world model in VLN-CE tasks, as conventional TTA methods alone yield limited improvements, underscoring the necessity of structured world modeling for online adaptation to novel tasks.

Additionally, the configurations of these TTA strategies for VLN are detailed as follows:
\begin{itemize}
	\item \textbf{Tent}~\cite{wang2020tent}. We adopt all hyperparameter settings as specified in Tent. Specifically, the optimizer is AdamW~\cite{Loshchilov2017DecoupledWD}, and for a batch size of 1, the learning rate is set to $0.001/64$. %The trainable parameters are all affine parameters of normalization layers.
	\item \textbf{NOTE}~\cite{gong2022note}. The hyperparameter configurations strictly follow those defined in NOTE. In particular, the soft-shrinkage width is set to $4$, and the EMA momentum is $0.01$. The optimization is performed using AdamW with a learning rate of $0.0001$.
	\item \textbf{SAR}~\cite{niu2023towards}. We adhere to the default hyperparameters in SAR. Specifically, the entropy constant $E_0$ (for reliable sample identification) is set to $0.4 \times \ln{1000}$, while the neighborhood size for sharpness-aware minimization is configured as $0.05$. For model recovery, the moving average factor is $0.9$, and the reset threshold is $0.2$. 
	\item \textbf{ViDA}~\cite{Liu2023ViDAHV}. The experimental setup follows the original ViDA configuration. Random augmentation compositions, including Gaussian noise and dropout, are incorporated. The AdamW optimizer, identical to that in Tent, is utilized. The threshold value is set to $0.2$, and the updating weight is $0.999$.
\end{itemize}

\begin{table}\centering
	\caption{Comparison with Representative TTA Strategies.}
	\label{tab:ablation-tta}
	\resizebox{0.99\linewidth}{!}{
		\begin{threeparttable} 
			\begin{tabular}{@{}l|ccccccc@{}}
				\toprule
				\multicolumn{1}{c|}{\multirow{2}{*}{\textbf{Methods}}} & \multicolumn{7}{c}{\textbf{R2R-CE Val Unseen}} \\ \cmidrule(l){2-8} 
				\multicolumn{1}{c|}{} & \cellcolor{red!25}TL $\downarrow$ & \cellcolor{red!25}NE $\downarrow$ & \cellcolor{gray!25}OSR & \cellcolor{gray!25}SR & \cellcolor{gray!25}SPL & \cellcolor{gray!25}NDTW & \cellcolor{gray!25}SDTW \\ \midrule
				\multicolumn{1}{l|}{Base Model} & 26.16 & 6.05 & 54.92 & 43.77 & 29.39 & 40.94 & 29.30 \\ \midrule
				\quad+ Tent~\cite{wang2020tent} & 28.56\scriptsize{$\pm$ 1.59} & 7.21\scriptsize{$\pm$ 1.01} & 52.13\scriptsize{$\pm$ 1.98} & 40.97\scriptsize{$\pm$ 1.77} & 27.46\scriptsize{$\pm$ 0.94} & 37.90\scriptsize{$\pm$ 1.53} & 27.65\scriptsize{$\pm$ 1.60} \\
				\quad+ NOTE~\cite{gong2022note} & 26.88\scriptsize{$\pm$ 1.82} & 6.71\scriptsize{$\pm$ 0.63} & 53.87\scriptsize{$\pm$ 1.71} & 42.85\scriptsize{$\pm$ 0.88} & 28.43\scriptsize{$\pm$ 0.56} & 39.02\scriptsize{$\pm$ 0.93} & 28.37\scriptsize{$\pm$ 0.88} \\
				\quad+ SAR~\cite{niu2023towards} & 27.15\scriptsize{$\pm$ 1.40} & 6.57\scriptsize{$\pm$ 0.83} & 53.50\scriptsize{$\pm$ 1.30} & 43.02\scriptsize{$\pm$ 0.91} & 27.98\scriptsize{$\pm$ 0.72} & 38.77\scriptsize{$\pm$ 0.95} & 27.92\scriptsize{$\pm$ 0.75} \\
				\quad+ ViDA~\cite{Liu2023ViDAHV} & 26.74\scriptsize{$\pm$ 1.26} & 6.88\scriptsize{$\pm$ 0.75} & 55.26\scriptsize{$\pm$ 0.98} & 43.58\scriptsize{$\pm$ 0.86} & 28.29\scriptsize{$\pm$ 0.53} & 40.89\scriptsize{$\pm$ 0.94} & 28.73\scriptsize{$\pm$ 0.56} \\
				\quad+ FSTTA~\cite{gao2024fast} & 28.25\scriptsize{$\pm$ 0.72} & 6.67\scriptsize{$\pm$ 0.34} & 55.41\scriptsize{$\pm$ 0.91} & 43.94\scriptsize{$\pm$ 0.32} & 29.63\scriptsize{$\pm$ 0.47} & 42.76\scriptsize{$\pm$ 0.65} & 29.34\scriptsize{$\pm$ 0.49} \\ \midrule
				\multicolumn{1}{l|}{\textbf{NavMorph}} & \textbf{22.54\scriptsize{$\pm$ 0.07}} & \textbf{5.75\scriptsize{$\pm$ 0.03}} & \textbf{56.88\scriptsize{$\pm$ 0.05}} & \textbf{47.91\scriptsize{$\pm$ 0.04}} & \textbf{33.22\scriptsize{$\pm$ 0.02}} & \textbf{44.86\scriptsize{$\pm$ 0.07}} & \textbf{32.73\scriptsize{$\pm$ 0.04}} \\ \bottomrule
			\end{tabular}
			\begin{tablenotes}    
				\footnotesize               
				\item Note: The reported values represent the mean results, with the standard deviation provided in a reduced font size. Best results are shown in bold.
			\end{tablenotes}           
		\end{threeparttable} 
	}
\end{table}

\noindent\textbf{Ablation Study of Action Embedding.}
In our proposed world model, action embedding $\boldsymbol{a}_t$ plays a crucial role in modeling state transitions and long-term predictive reasoning within RSSM, enabling the model to generate plausible future trajectories based on past observations and actions (\underline{\textit{Section 3.1 in main paper}}).
%By mapping discrete actions into a continuous feature space, action embedding provides a structured representation of agent movements.
Some readers may wonder whether the model’s performance improvement arises from learning an action-state mapping through action embedding.
To further investigate this, we conduct an ablation study to assess its role in latent representation learning. Specifically, we introduce a variant, `Base-AE', which retains the same backbone as the baseline model (VLN-3DFF~\cite{wang2024sim}) without world modeling but includes an additional action embedding input.

As shown in Table~\ref{tab:action}, the results indicate only marginal differences (OSR: 54.9$\to$55.1, SR: 43.8$\to$43.4, SPL: 29.4$\to$29.3), indicating that explicitly encoding actions has a negligible effect on performance. This finding highlights that the observed improvements in our method stem primarily from the self-evolving world model’s ability to model environmental dynamics, rather than the mere inclusion of action embedding.

\begin{table}\centering
	\caption{Ablation Study of Action Embedding.}
	\label{tab:action}
	\resizebox{0.98\columnwidth}{!}{
		\begin{threeparttable}
			\begin{tabular}{@{}c|ccccccc@{}}
				\toprule
				\multirow{2}{*}{\textbf{Methods}} & \multicolumn{7}{c}{\textbf{R2R-CE Val Unseen}} \\ \cmidrule(l){2-8} 
				& \cellcolor{red!25}TL $\downarrow$ & \cellcolor{red!25}NE $\downarrow$ & \cellcolor{gray!25}OSR & \cellcolor{gray!25}SR & \cellcolor{gray!25}SPL & \cellcolor{gray!25}NDTW & \cellcolor{gray!25}SDTW \\ \midrule
				Base Model & 26.16 & 6.05 & 54.92 & 43.77 & 29.39 & 40.94 & 29.30 \\ \midrule
				Base-AE & 26.01 & 6.09 & 55.1 & 43.4 & 29.3 & 41.34 & 29.42 \\
				NavMorph & \textbf{22.54} & \textbf{5.75} & \textbf{56.88} & \textbf{47.91} & \textbf{33.22} & \textbf{44.86} & \textbf{32.73} \\ \bottomrule
			\end{tabular}
			\begin{tablenotes}    
				\footnotesize               
				\item Note: `Base Model' denotes the chosen baseline under monocular setting, VLN-3DFF. `Base-AE' denotes the baseline incorporating action information into the input without the world model. Best results are shown in bold.
			\end{tablenotes}           
		\end{threeparttable} 
	}
\end{table}

%Instead, Navmorph effectively learns to capture environmental dynamics which aids for navigation.

%It is noteworthy that some variants of our model exhibit a slight decrease in the OSR metric, which can be attributed to the inherent trade-off between exploration and efficiency in VLN tasks. While our method emphasizes more efficient navigational inference by generating direct paths to the stop point, this approach can occasionally result in fewer points falling within the 3-meter threshold used by OSR.

\noindent\textbf{Computational Analysis.} \underline{\textit{Table 4 in main paper}} demonstrates that NavMorph maintains comparable inference efficiency to the base model, with an average episode time of 21.22s \textit{vs} 20.53s, while achieving nearly 4\% improvements in both SR and SPL. 
In terms of parameter overhead, our CEM introduces only a marginal increase—adding 2.30M parameters compared to the base model's 228.96M, accounting for merely 1.0\% of the total model size. %During inference on the R2R-CE validation unseen split, each episode consists of approximately 9.23 timesteps, with one scene feature vector $v_m$ generated per step.
Given the consistent performance improvements, the computational and memory overheads are lightweight and acceptable.

\begin{figure*}[t!]
	\centering
	\includegraphics[width=0.98\linewidth]{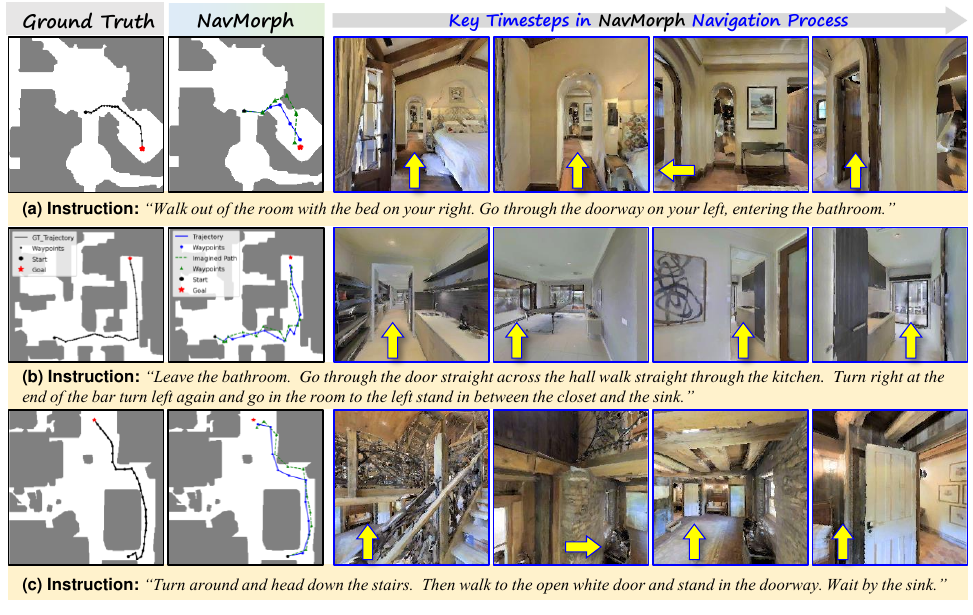}
	\caption{Qualitative results of NavMorph on the R2R-CE dataset are presented, showcasing a comparison between ground truth paths (GT\underline{~}Trajectory), NavMorph's executed navigation routes (Trajectory), and the predictive paths generated by the Foresight Action Planner (Imagined Path). These visualizations highlight NavMorph's ability to perform effective navigation. Additionally, key input observations at critical timesteps during NavMorph's navigation are provided to illustrate its decision-making process.}
	\label{fig:vis}
\end{figure*}

\subsection{Extensive Discussion}\label{sesc:discussion}
While our method shares similarities with test-time adaptation (TTA), NavMorph fundamentally extends beyond this paradigm. Traditional TTA typically applies gradient-based parameter updates during inference for static classification tasks. In contrast, our approach incorporates a contextual evolution mechanism (CEM) within the RSSM framework, explicitly modeling environment state transitions during both training and inference phases. This mechanism enables the agent to adapt proactively to dynamic environments—not just during inference—by selectively integrating new scene observations while retaining historically relevant information.

At the core of this design lies the Contextual Evolution Memory (CEM), which enhances long-term reasoning by dynamically maintaining latent scene representations. Rather than accumulating all past experiences, CEM performs top-K scene retrieval based on visual similarity, maintaining memory entries that are most pertinent to the current context. This suppresses noisy or outdated trajectories and enhances the agent’s ability to infer plausible future transitions, particularly in out-of-distribution or evolving scenes, as often encountered in VLN-CE tasks.
We provide ablations (Table~\ref{tab:ablation-se}) comparing variants of NavMorph with and without online evolution (NavMorph w/o SE*), confirming that the model remains effective even without runtime adaptation (SR: Base Model 43.77 → NavMorph w/o SE* 46.87 → NavMorph 47.91).

Importantly, our contribution lies not in the memory design itself, but in showing that adaptive evolution—when tightly integrated into a world model—can effectively improve navigation performance in VLN-CE literature.

\section{Qualitative Analysis}\label{sec:quali}
%\textcolor{red}{Visualization for VLN-CE tasks. }
%To analyze the predictive capabilities of NavMorph, we visualize the trajectories generated by our Foresight Action Planner and compare them with actual execution paths and ground truth actions.  Figure~\ref{fig:vis} illustrates predicted and actual navigation coordinates across diverse scenarios. The alignment between predicted and executed trajectories demonstrates our model effectively captures environmental dynamics and aids foresight planning.We observe that in simple scenarios with clear navigation paths (Figure~\ref{fig:vis}(a)), the predicted trajectories closely match the actual execution, indicating robust state modeling. In more complex environments across multiple room types (Figure~\ref{fig:vis}(b)), while slight deviations occur at decision points, NavMorph maintains overall trajectory consistency. Even in challenging cases with partial occlusions (Figure~\ref{fig:vis}(c)), where descending stairs creates visual discontinuities and partial occlusions of the target path, the model demonstrates resilient prediction capabilities by leveraging historical context through CEM. These visualization results validate that our world model effectively captures environmental dynamics and spatial relationships, enabling reliable foresight planning during navigation. 

To evaluate the predictive performance of NavMorph, we conduct qualitative analysis by comparing trajectories generated through our Foresight Action Planner with executed paths and ground truth sequences. 
Since our world model encodes high-level features instead of raw images, direct visualization of latent states remains non-trivial. Therefore, to effectively illustrate NavMorph’s reasoning process, we resort to trajectory-based evaluations, where the predicted and executed navigation sequences serve as an implicit reflection of the model’s latent space dynamics.
Figure~\ref{fig:vis} presents trajectory visualizations across diverse navigational scenarios, where the coherence between predicted and executed paths demonstrates the model's capacity for environmental dynamics modeling and anticipatory planning. Furthermore, we visualize key observational inputs at critical navigation timesteps to provide insights into NavMorph's decision-making process.

%We observe that in simple scenarios with clear navigation paths (Figure~\ref{fig:vis}(a)), the predicted trajectories closely match the actual execution, indicating robust state modeling. For more complex environments involving multiple room transitions (Figure~\ref{fig:vis}(b)), NavMorph maintains trajectory consistency despite minor deviations at critical decision points. Notably, in challenging multi-level scenarios (Figure~\ref{fig:vis}(c)), where descending stairs introduces visual discontinuities and occlusions, the model preserves prediction accuracy. 

%It is worth noting that in some cases, the imagined path generated by the foresight action planner (\textit{i.e.}, the green dashed trajectory) appears to cross through obstacles (e.g., walls or furniture). This limitation arises because the world model's imaginative rollouts are not guided by immediate feedback, such as navigation rewards, to penalize unrealistic paths that intersect with obstacles. Despite this, the foresight action planner serves as a valuable approximation, providing target-oriented trajectories that the world-aware navigator can refine in real-time to accommodate environmental constraints. These comprehensive visualizations substantiate our world model's capability in capturing both environmental dynamics and spatial-temporal relationships, enabling plausible predictive planning during navigation tasks.

We observe that in simple scenarios with clear navigation paths (Figure~\ref{fig:vis}(a)), the predicted trajectories closely align with the actual execution, demonstrating robust state modeling capabilities. For complex environments involving multiple room transitions (Figure~\ref{fig:vis}(b)), NavMorph maintains trajectory consistency with only minor deviations at critical decision points. Notably, in challenging multi-level scenarios (Figure~\ref{fig:vis}(c)), where descending stairs introduces visual discontinuities and occlusions, the model exhibits resilient prediction performance.

Notably, the imagined paths predicted by the foresight action planner (indicated by green dashed trajectories) occasionally traverse through physical obstacles such as walls or furniture. This limitation stems from the world model's imaginative rollouts lacking immediate environmental feedback, particularly navigation rewards that would typically penalize obstacle intersections. Nevertheless, the foresight action planner proves effective as an approximation mechanism, generating target-oriented trajectories that the world-aware navigator can dynamically adjust during execution to satisfy environmental constraints. These visualization results validate our world model's proficiency in capturing environmental dynamics and spatial-temporal relationships, facilitating effective predictive planning in navigation tasks.

\end{document}